\documentclass{article}

\usepackage{iclr2027_conference,times}
\usepackage[T1]{fontenc}
\usepackage{adjustbox,needspace,etoolbox}
\usepackage{amsmath,amssymb,amsthm}
\usepackage{array}
\usepackage{booktabs}
\usepackage{enumitem}
\usepackage{float}
\usepackage[colorlinks=true,citecolor=blue,linkcolor=blue,urlcolor=blue]{hyperref}
\usepackage{listings}
\usepackage{graphicx}
\usepackage{pgfplots}
\usepackage[most]{tcolorbox}
\usepackage{type1cm}
\usepackage{xspace}
\usepgfplotslibrary{groupplots}
\pgfplotsset{compat=1.17}
\newtcblisting{promptbox}[1]{%
  breakable, enhanced, listing only,
  colback=black!3, colframe=black!45, coltitle=black,
  fonttitle=\bfseries\footnotesize, title={#1},
  boxrule=0.5pt, arc=2pt, left=6pt, right=6pt, top=3pt, bottom=3pt,
  listing options={basicstyle=\ttfamily\scriptsize, breaklines=true, breakindent=1em,
    columns=fullflexible, keepspaces=true, extendedchars=true, literate={—}{{-{}-}}1}%
}

\setlist[itemize]{leftmargin=1.5em,itemsep=0.15em,topsep=0.15em}
\setlist[enumerate]{leftmargin=1.7em,itemsep=0.15em,topsep=0.15em}
\title{Measuring Intelligence Beyond Human Scale}

\iclrfinalcopy
\makeatletter
\patchcmd{\@maketitle}
  {\lhead{Published as a conference paper at ICLR 2027}}
  {\lhead{}}
  {}
  {\PackageError{separank}{Could not remove conference publication header}{Check the ICLR style before compiling.}}
\patchcmd{\@maketitle}
  {\lhead{Under review as a conference paper at ICLR 2027}}
  {\lhead{}}
  {}
  {\PackageError{separank}{Could not remove conference review header}{Check the ICLR style before compiling.}}
\makeatother

\author{%
\parbox{\textwidth}{\centering
Jerry Han\thanks{Equal contribution.}\quad
Rafael Moschopoulos\footnotemark[1]\quad Ella Colby\\[3pt]
Vishrut Goyal\quad Andrew Tu\quad Kia Ghods\\[3pt]
Shirley Yu\quad Mark Braverman\quad Elad Hazan\\[6pt]
{\normalfont Princeton AI Alignment and Safety \\ \url{https://alignment.dais.princeton.edu/}  \thanks{We gratefully acknowledge contributions by Iris Yan and Connor Brown.}}%
}}

\begin{document}
\maketitle

\begin{abstract}
How can we measure intelligence beyond human capability?
Human-authored benchmarks saturate easily. When models exceed human capability, human examiners might have trouble coming up with both challenging and verifiable tasks. We argue that this difficulty is an inherent property of absolute-scale evaluation. We propose a new paradigm based on relative measurements in which models generate public challenges that separate other systems. We develop a novel protocol that naturally scales with agent capabilities, requires no human-authored questions, and is also judge-free. We instantiate this framework across verifiable and open-ended, non-verifiable domains, illustrating a route by which model-generated evaluation could continue to measure systems as they approach and exceed the human frontier.
\end{abstract}

\section{Introduction}

How should intelligence be measured? The question is older than the study of artificial intelligence.

The modern scientific study of intelligence began with \citet{spearman1904}, who observed that performance on different cognitive tests is positively correlated and modeled this with a single latent factor, general intelligence ($g$) (Section~\ref{sec:spearman}).

Most AI benchmarks share the examiner-designed format of classical testing, although few fit latent-trait models such as factor analysis or item response theory. Systems are evaluated on tasks supplied by humans or in environments fixed in advance, such as game environments and broad academic benchmarks \citep{hendrycks2021mmlu,srivastava2022bigbench,liang2022helm,chollet2019arc,white2024livebench,glazer2024frontiermath,phan2025hle,kiela2021dynabench}.

Many of these benchmarks are now saturated, or close to being saturated, by frontier AI models. This reduces their ability to discriminate between models of different capabilities and requires harder benchmarks, which become considerably more difficult to construct as AI models become more capable than human test writers.

\paragraph{Relative intelligence measurement.}

A different starting point is relative measurement. Turing's imitation game measures machine intelligence by comparison with human behavior before a human judge, rather than by an absolute score on a fixed test \citep{turing1950}. Other current methods pit 2 models against each other in a competitive game, and resolve the result with a judge \citep{zheng2023judge,chiang2024arena}. But these setups often still rely on human-authored questions or human judges to choose the preferred or winning response. A notable exception is MathDuels \citep{xu2026mathduels}, which uses model-generated mathematical challenges, solved across the model population, as an evaluation primitive.

However, models can do well in these setups in ways that do not translate to actual model capabilities. For example, models can propose questions that are contingent on private information, or cryptographically hard questions that are easy to verify but much harder to solve (Section~\ref{subsection:pairwise_caveats}).

\paragraph{A different approach to measuring intelligence.}

We propose a model-generated and separative approach to intelligence evaluation, which we call \emph{adversarial psychometrics}. While classical psychometrics measures intelligence by how well subjects solve examiner-designed tasks, adversarial psychometrics evaluates a model's ability to discover capability boundaries among other systems. In classical psychometrics, a good test item is one that discriminates between subjects of different ability, and SepaRank rewards proposers for writing exactly such items. We hypothesize that a model displays an important component of intelligence by constructing challenges that separate a population of solvers.

Consider a question that is proposed publicly to several models. If they all respond with approximately the same confidence level, the question tells us very little. However, if the question induces some models to be extremely confident in their answers while others are not, then it is an informative question that discriminates between various models. Therefore, our key insight is to evaluate models on their capacity to discover such capability gaps among other models. 

\paragraph{Paper overview.}
We begin by reviewing Spearman's general intelligence $g$ (Section~\ref{sec:spearman}), which motivates measuring intelligence through agreement across many tests and provides the $g$-loading we later use to compare SepaRank with existing benchmarks. We then demonstrate the natural limitations of pairwise challenge protocols, which are fundamentally susceptible to private-information and trapdoor attacks. This motivates a new one-to-many formulation, which we develop further in the second part of this paper. We present our separative psychometric framework and explain its scoring rule, adaptive weighting and judge-free properties. Finally, we verify the robustness of this framework empirically across both verifiable and non-verifiable domains. We demonstrate that our proposed framework for measuring model capability correlates well with existing benchmarks of model capabilities.

\subsection{Relation to existing work}

Psychometric and adaptive-testing methods estimate ability from answers to test questions written and given by human examiners \citep{spearman1904,rasch1960,lord1980,vanderlinden2010}.

Bradley--Terry and Elo rate participants based on the win/loss outcomes of matches between participants. TrueSkill generalizes this to handle matches with more than 2 players or with teams, and also keeps track of the uncertainty in the rating \citep{bradley1952,elo1978,trueskill2007}. Modern AI evaluation has focused on benchmark construction, contamination resistance, frontier-level difficulty, dynamic task collection, and pairwise comparison, including MMLU, BIG-Bench, HELM, ARC, LiveBench, FrontierMath, Humanity's Last Exam, Dynabench, MT-Bench, and Chatbot Arena \citep{hendrycks2021mmlu,srivastava2022bigbench,liang2022helm,chollet2019arc,white2024livebench,glazer2024frontiermath,phan2025hle,kiela2021dynabench,zheng2023judge,chiang2024arena}.

In setups like Language-Model-as-an-Examiner, Auto-Arena, and PeerRank, models author questions and evaluate the responses, while retaining semantic judgements as the scoring mechanism \citep{review_bai2023examiner,review_zhao2025autoarena,review_margalit2026peerrank}. LivingArena looks for questions that an opponent is likely to struggle with and uses 3 other models to judge the answers \citep{review_chen2026livingarena}. MathDuels \citep{xu2026mathduels} is particularly related to our work. It proposes a mathematical self-play benchmark in which models act as both problem posers and solvers. 

Our work generalizes the idea of measuring the ability of models to generate interesting tasks. Our approach differs in several important aspects. We consider arbitrary general domains of tasks beyond mathematics. We also develop a psychometric framework based on separating a population of solvers, who are scored against proposer-committed answers. Finally, we consider protocols that do not require a semantic judge.

\paragraph{Relation to peer prediction.}

Peer prediction studies how to elicit truthful reports when no ground truth is available to score them against. The peer-prediction method of \citet{miller2005peer} scores each report by how well it predicts another agent's report under a proper scoring rule. The Bayesian truth serum \citep{prelec2004bts} rewards answers that are more common than respondents collectively predict. Such mechanisms admit uninformative equilibria in which agents can coordinate, so later multi-task mechanisms reward agreement only beyond what is expected by chance \citep{dasgupta2013crowdsourced,shnayder2016informed}. SepaRank faces a similar problem in non-verifiable domains, where solvers are scored against the proposer's private answer commitment rather than an external truth, similarly to how a peer-prediction report is scored against a peer's report. Our goal differes however in that we seek to measure capability rather than elicit truthful reports, and the proposer is rewarded for disagreement among solvers rather than agreement. (see Section~\ref{subsec:resolution-rules}).

\paragraph{Relation to scalable oversight.}

How can people supervise an AI model if it can solve problems they cannot? This is a key problem studied in scalable oversight \citep{amodei2016,christiano2018}. A notable proposed mechanism is AI safety via debate, in which models argue opposing answers before a weaker judge \citep{irving2018}. Apart from this, there is also weak-to-strong generalization, which proposes having a weak supervisor to elicit the full capabilities of a much stronger model \citep{burns2023}. And finally, there are sandwiching-style experiments, in which non-experts use model assistance on tasks that are then evaluated against expert-defined answers \citep{bowman2022}. 

Our work shares the same motivation as many of these ideas in the field of scalable oversight. Given that human graders might eventually be unable to author or verify sufficiently difficult and discriminative tasks, we need to move to a paradigm of comparative measurement. 

\paragraph{Relation to interactive proofs and complexity theory.}

Our work is also closely related to multi-prover interactive proofs (MIPs). In an MIP, a polynomial-time verifier communicates with several computationally unbounded provers that cannot communicate with each other. With 2 or more provers, $\mathsf{MIP} = \mathsf{NEXP}$ \citep{babaiMIP1991}, so a weak verifier can check claims that would take it exponentially more time to verify on its own.

Our protocol is not an MIP in the formal sense, as we do not provide completeness or soundness guarantees for a fixed language, and we do not assume a polynomial-time verifier. However, we share the same underlying principle with MIPs that comparing several independent agents reveals information that querying any single agent cannot. While MIPs use cross-prover consistency to verify mathematical claims, we use cross-model disagreement to measure capability boundaries of models.

\definecolor{spPos}{HTML}{2166AC}

\definecolor{spNeg}{HTML}{B2182B}

\newcommand{\spfont}{\scriptsize}

\newcommand{\spcell}[4]{%
  \pgfmathsetmacro{\spt}{min(85,85*abs(#3)/#4)}%
  \ifdim #3pt<0pt \colorlet{spfill}{spNeg!\spt!white}\else\colorlet{spfill}{spPos!\spt!white}\fi
  \fill[spfill] (#1-1,1-#2) rectangle (#1,-#2);%
  \ifdim \spt pt>50pt \def\sptc{white}\else\def\sptc{black}\fi
  \node[text=\sptc, font=\spfont] at (#1-0.5,0.5-#2) {#3};}

\newcommand{\spdiag}[2]{%
  \fill[black!7] (#1-1,1-#1) rectangle (#1,-#1);%
  \node[text=black!45, font=\scriptsize] at (#1-0.5,0.5-#1) {#2};}

\newcommand{\spmatrix}[2]{
  \foreach \row [count=\r] in {#1}
    \foreach \v [count=\c] in \row { \ifnum\r=\c\relax\else \spcell{\c}{\r}{\v}{1}\fi }
  \foreach \d [count=\c] in {#2} { \spdiag{\c}{\d} }
  \draw[black!45] (0,0) rectangle (6,-6);}

\section{Background on Spearman's general intelligence \texorpdfstring{$g$}{g}}
\label{sec:spearman}

In the early 1900s, experimental psychologists were intrigued by the question of measuring human intelligence. Spearman followed Galton's theories of intelligence but noted that Galton surprisingly failed to find associations between different performance metrics and common indicators of intelligence. 

\begin{figure}[!htbp]
\centering
\begin{minipage}{0.6\linewidth}\centering
\includegraphics[width=\linewidth]{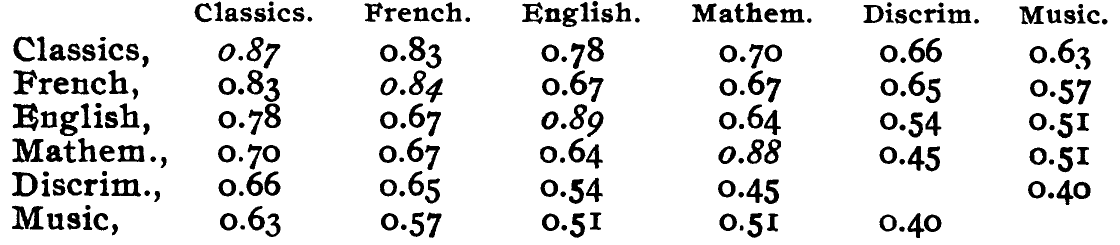}\par\smallskip
{\small (a) Correlations between tests (p.~275)}
\end{minipage}\par\medskip
\begin{minipage}{0.6\linewidth}\centering
\includegraphics[width=\linewidth]{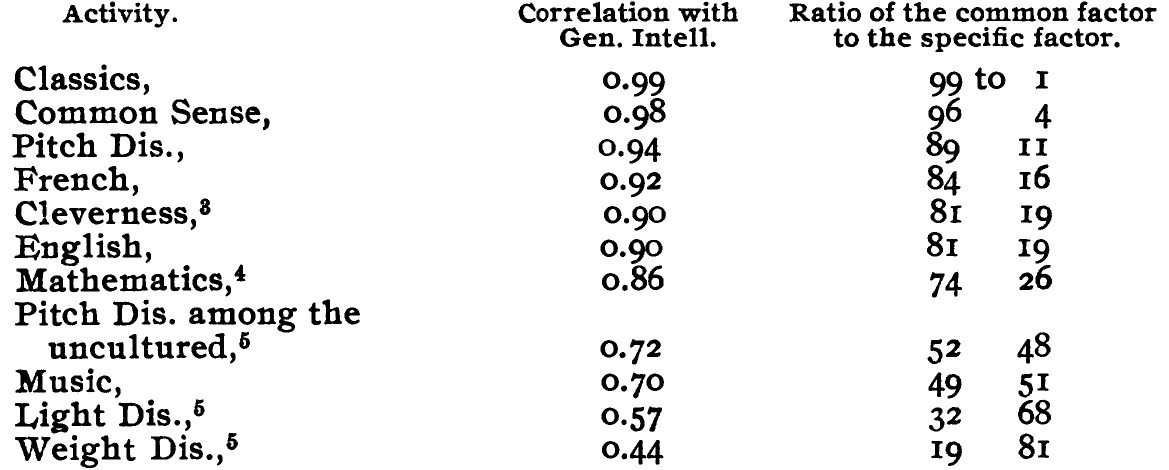}\par\smallskip
{\small (b) Correlations with $g$ (p.~276)}
\end{minipage}
\caption{Spearman's 1904 tables \citep{spearman1904}. (a) Correlations among the 6 measures for the 22 music pupils of a preparatory school (b) Spearman's estimates of each activity's correlation with $g$, pooled across his experiments.}
\label{fig:spearman-original}
\end{figure}

\paragraph{Spearman's experiment.} Spearman set out ``to determine this Intelligence in a definite objective manner, and to discover means of precisely measuring it'' \citep[p.~206]{spearman1904}. His clearest data came from a preparatory school for boys. He ranked the pupils by their results in the regular school examinations in Classics, French, English and Mathematics. He also tested their discrimination of pitch and obtained a ranking of musical ability for the 22 boys taking music lessons \citep[pp.~248--250, 270]{spearman1904}.Figure~\ref{fig:spearman-original}(a) reproduces his table of correlations among these 6 measures for those 22 boys \citep[pp.~275, 291]{spearman1904}.

\paragraph{The positive manifold.} Spearman observed that every correlation in the table is positive. A boy who ranks high in one subject tends to rank high in all of them, and even to hear finer differences in pitch. The entries are also orderly, falling steadily ``as the eye travels either to the right or downwards,'' a pattern Spearman called the \emph{hierarchy} \citep[p.~275]{spearman1904}. He concluded that ``all branches of intellectual activity have in common one fundamental function,'' plus a specific element unique to each \citep[p.~284]{spearman1904}. He called it General Intelligence, later written $g$.

\paragraph{The one-factor model.} In mathematical terms, person $i$'s score on test $k$ is $s_{ik}=\alpha_k+\lambda_k g_i+\varepsilon_{ik}$, with one ability $g_i$ per person, one loading $\lambda_k$ per test, and a test-specific remainder $\varepsilon_{ik}$. The person-by-test score matrix is therefore approximately rank one, $X\approx g\lambda^\top$. If the specific parts are uncorrelated, the correlation between 2 tests $k\neq k'$ is a product, $r_{kk'}=\ell_k\ell_{k'}$, where $\ell_k=\operatorname{corr}(\text{test } k, g)$ is the \emph{$g$-loading} of test $k$. The correlation matrix is then rank one off its diagonal, and sorting tests by $\ell_k$ yields the hierarchy.

\paragraph{Correlation with general intelligence.} Figure~\ref{fig:spearman-factorization} shows that Spearman's table is approximately rank one. It is the product of a single vector with its own transpose, $\ell\ell^\top$ with $\ell=(0.96,0.88,0.80,0.75,0.67,0.65)$, plus a small residual. Each $\ell_k$ estimates the correlation of test $k$ with $g$. In this rank-one fit, Classics ($0.96$) and French ($0.88$) load most strongly on $g$, while pitch ($0.67$) and music ($0.65$) load least.

\begin{figure}[!htbp]
\centering
\begin{adjustbox}{max width=\linewidth}
\begin{tikzpicture}[x=0.62cm, y=0.43cm]
  \begin{scope}
    \spmatrix{{0,.83,.78,.70,.66,.63},{.83,0,.67,.67,.65,.57},{.78,.67,0,.64,.54,.51},
              {.70,.67,.64,0,.45,.51},{.66,.65,.54,.45,0,.40},{.63,.57,.51,.51,.40,0}}
             {--,--,--,--,--,--}
    \foreach \nm [count=\r] in {Classics,French,English,Maths,Pitch,Music}
      \node[anchor=east, font=\scriptsize] at (-0.1,0.5-\r) {\nm};
    \foreach \nm [count=\c] in {Cla,Fre,Eng,Mat,Pit,Mus}
      \node[font=\scriptsize] at (\c-0.5,0.45) {\nm};
    \node[font=\small, align=center] at (3,-7.0) {observed $r_{kk'}$\\[-1pt]{\scriptsize Spearman (1904), p.~275}};
  \end{scope}
  \node[font=\Large] at (7.0,-3) {$\approx$};
  \begin{scope}[shift={(9.4,0)}]
    \foreach \v [count=\c] in {.96,.88,.80,.75,.67,.65} {
      \begin{scope}[shift={(0,1.3)}] \spcell{\c}{1}{\v}{1} \end{scope} }
    \draw[black!70, thick] (0,0.3) rectangle (6,1.3);
    \node[anchor=west, font=\small] at (6.1,0.8) {$\ell^{\top}$};
    \foreach \v [count=\r] in {.96,.88,.80,.75,.67,.65} {
      \begin{scope}[shift={(-1.3,0)}] \spcell{1}{\r}{\v}{1} \end{scope} }
    \draw[black!70, thick] (-1.3,0) rectangle (-0.3,-6);
    \node[font=\small] at (-0.8,-6.55) {$\ell$};
    \node[font=\large] at (-0.8,0.8) {$\times$};
    \spmatrix{{0,.84,.77,.72,.64,.62},{.84,0,.71,.66,.59,.57},{.77,.71,0,.60,.54,.52},
              {.72,.66,.60,0,.50,.48},{.64,.59,.54,.50,0,.43},{.62,.57,.52,.48,.43,0}}
             {.92,.78,.64,.56,.45,.42}
    \node[font=\small, align=center] at (3,-7.0) {rank-one fit $\ell_k\ell_{k'}$\\[-1pt]{\scriptsize $\ell_k=\operatorname{corr}(\text{test } k,\, g)$; cell $=$ row $\times$ column}};
  \end{scope}
  \node[font=\Large] at (16.4,-3) {$+$};
  \begin{scope}[shift={(17.4,0)}]
    \renewcommand{\spfont}{\fontsize{6}{7}\selectfont}
    \spmatrix{{0,-.01,.01,-.02,.02,.01},{-.01,0,-.04,.01,.06,.00},{.01,-.04,0,.04,.00,-.01},
              {-.02,.01,.04,0,-.05,.03},{.02,.06,.00,-.05,0,-.03},{.01,.00,-.01,.03,-.03,0}}
             {,,,,,}
    \node[font=\small, align=center] at (3,-7.0) {residual\\[-1pt]{\scriptsize all $|{\cdot}|\le 0.06$}};
  \end{scope}
\end{tikzpicture}
\end{adjustbox}
\caption{Spearman's correlations $\approx$ rank-one $\ell\ell^\top$ $+$ small residual; $\ell_k=\operatorname{corr}(\text{test } k, g)$.}
\label{fig:spearman-factorization}
\end{figure}

In Section~\ref{subsec:external-benchmarks}, we treat SepaRank and the 10 benchmarks of the Artificial Analysis Intelligence Index as tests in a single battery and estimate the $g$-loading $\ell_k$ of each with a leave-one-out proxy for $g$.

\section{Warm-up: pairwise intelligence tournaments}

\label{sec:caveats}

A natural first attempt is to compare 2 models at a time. Consider models $i$ and $j$. We ask $i$ to propose a challenge that it can solve but $j$ cannot solve. The challenge should be publicly specified and objectively resolvable (any sufficiently intelligent or capable agent should be able to read the question and verify the answer). Both models should be accorded the same resource budget, and $i$ is awarded a victory if it successfully poses a question that it can solve and $j$ cannot. 

After repeating this with many different pairs over many matches, the outcomes of these matches define a directed graph, where an edge $i \to j$ means that $i$ found a question it could answer correctly but $j$ could not. We can use Bradley--Terry, Elo, or TrueSkill to transform this graph into ratings. For example, Bradley--Terry gives each model $i$ a score $\theta_i$. The difference between the 2 scores determines how likely one model is to beat the other,
\[
\Pr(i \text{ beats } j) = \sigma(\theta_i-\theta_j),
\]
where $\sigma$ is the logistic function.
We can also use the Elo rating system to update each model's rating after every match.

As models improve, they may come up with questions that weaker models cannot answer. This framework also allows us to continue adding new models and updating the ratings based on the results of additional games played. That being said, there are some fundamental flaws with this pairwise paradigm. 

\subsection{Caveats of pairwise protocols} \label{subsection:pairwise_caveats}

The first problem is that a model only needs to beat one opponent. It could choose a question that the other model has no way of answering. For example, it could ask, ``which bit am I thinking of?'' Another problematic construction is for a proposer to generate 2 large primes, publish their product, and ask for its factorization. This task is much harder for the solver, so under the same resource budget it would likely be unsolvable for the solver, whereas the proposer can solve it but only because it inserted the trapdoor during the construction of the problem.

In both cases, the proposer possesses some form of private information that is unavailable to the solver. In the first case it is a private state, and in the second case it is an algorithmic trapdoor. Such a challenge which stumps the solver does not necessarily reflect a capability difference between the proposer and solver, since even a massively more capable solver would not have been able to solve it. Indeed, this difficulty arises from the pairwise reward, which rewards the proposer for the asymmetry between proposer and solver, rather than for revealing capability differences between the solvers. 

Another problem that arises with pairwise protocols is adjudication. Even if we manage to exclude such questions that rely on private information, the protocol still necessitates an external adjudicator to determine if a challenge is admissible, and whether the proposed solution is correct. This might be doable for verifiable domains like math or coding, but the reliance on external adjudication limits scalability to other more general non-verifiable domains. 

Our key observation is that such private information challenges only succeed because the proposer needs to defeat a single opponent. If the same challenge is posed simultaneously to many independently trained systems, the reward depends on population disagreement rather than success against one solver. In an appropriately designed mechanism, shared uncertainty among players is indicative of an uninformative challenge and should yield no separation reward. This motivates the one-to-many protocol developed next.

\section{The SepaRank Protocol}
\label{sec:protocol}

We call our novel ranking protocol SepaRank. Its basic description is as follows. A proposer submits a binary question with answers $0$ and $1$.\footnote{Binary questions are without loss of generality. A question that expects an answer of finite length can be reduced to a sequence of binary questions (e.g., by asking for successive bits of the answer or by using a binary search over the answer space). Since our objective is to measure separation power without unnecessary complications, we restrict our protocol to binary questions.} Given a question, each solver responds with a number in $[0, 1]$ indicating the solver's confidence that the answer committed by the question proposer is $1$. The proposer is rewarded for inducing high variance in the values $\{p_j\}$ across solvers $j$.
In general, we do not thematically constrain the content of the question, although we prompt the proposers to pose a binary question that is objectively resolvable. In reality, the question may be factual, mathematical, philosophical, subjective, future-dependent, or otherwise difficult to resolve. In Section~\ref{sec:experiments}, we also run a SepaRank version where questions must be Python programs, allowing us to know the real answer to the question. We call this version \emph{code-constrained}, and the standard version \emph{open-ended}.

Rewarding a question by the confidence variance it induces changes the incentive of proposers compared to pairwise challenge protocols. The question ``which bit am I thinking of?'' no longer succeeds, since a rational solver has no evidence and should report $p_j=\frac{1}{2}$, so the variance across $\{p_j\}$ is small. Similarly, a trapdoored factoring problem (factorization of a product of 2 large private primes) does not help the proposer, even though the proposer is not required to solve the problem and commit truthfully. If the problem is too hard, calibrated solvers recognize the objective difficulty of answering correctly and again report near $\frac{1}{2}$. The proposer succeeds only by finding questions that genuinely separate the audience. We conjecture that this discourages some uninformative challenges.

We use variance as the proposer objective for simplicity. Other separation objectives are also possible, such as the empirical entropy of the $\{p_j\}$ distribution or a weighted Jensen--Shannon divergence, which measures how much information a report carries about which solver produced it.

\subsection{Protocol}

In the SepaRank protocol, each of $N$ models participates as both a proposer and a solver.
The protocol consists of $R$ rounds, each of which consists of $N$ matches. In round $r \in [R]$, match $i \in [N]$,
model $i$ proposes one binary question $q^{(r)}_i$ and privately commits to an answer
\(y^{(r)}_i\in\{0,1\}\), and models $[N] \setminus \{i\}$ are prompted to answer that question. All $N$ matches during round $r$ use the full information of the game up to round $r-1$. Each model is assigned a stable pseudonym which identifies its questions, answers, and scores, all visible in the transcript.\footnote{In practice, we truncate the protocol transcript to the past 5 rounds for computational feasibility. However, model scores and standings are computed from round $1$ through $r-1$.} Participants in round $r$ cannot see each other's answers during that round, and may only access information up to round $r-1$. Only after round $r$ are scores, weights, and transcript updated. That way, matches can run in parallel and finish faster.

During round $r$, each solver \(j \in [N] \setminus \{i\}\) reports a
confidence \(p^{(r, i)}_j\in[0,1]\) that the committed answer $y^{(r)}_i$ is \(1\).
Reporting \(1/2\) expresses equal uncertainty between the 2 answers,
while values near \(0\) or \(1\) express confidence.

The proposer is rewarded for separating the solvers' beliefs. For one question, let $p^{(r, i)} = (p^{(r, i)}_j)_{j \in [N] \setminus \{i\}}$ be the solver reports and let
\( w^{(r)}_{-i} = (w^{(r)}_{j})_{j \in [N] \setminus \{i\}} \) be their round-$r$ weights, to be defined in Section~\ref{subsec:adaptive-weighting}.
Since the weights $w^{(r)}$ sum to $1$ over all $N$ models, we renormalize them over the solvers,
\(\tilde w^{(r, i)}_j = w^{(r)}_j \big/ \sum_{k \in [N] \setminus \{i\}} w^{(r)}_k\).
The proposer reward is $P^{(r)}_i$, defined by
\[
\mu^{(r)}_i=\sum_{j \in [N] \setminus \{i\}}\tilde w_j^{(r, i)} p^{(r, i)}_j,
\qquad
P^{(r)}_i=4\sum_{j \in [N] \setminus \{i\}} \tilde w_j^{(r, i)}(p^{(r, i)}_j-\mu^{(r)}_i)^2.
\]
This is the weighted population variance of the reported probabilities.
The factor \(4\) is used so that the reward is in \([0,1]\). It is $0$ when all
solvers report the same probability, and reaches $1$ when half the
weight reports \(0\) and the other half reports \(1\).
In practice, if a solver fails to report properly, they are not considered in this computation and the successful solvers' weights are appropriately normalized.
A proposer whose challenge is invalid (i.e., malformed, or a program that fails validation) instead receives reward $P^{(r)}_i=-1$, which counts toward its mean proposer score; the $[0,1]$ range above therefore applies to valid challenges only.

Solvers are scored separately against the proposer's committed answer by
\[
L^{(r, i)}_j=(p_j^{(r, i)}-y^{(r)}_i)^2.
\]
This proper scoring rule is called \emph{Brier loss} \citep{review_brier1950,review_gneiting2007}, and is smallest for an accurate, confident report and
largest for a confidently incorrect one.
For a commitment $y$ and a report $p$, letting $\pi=\Pr(y=1)$ denote the solver's belief, the expected loss is
\[
\mathbb{E}[L]=\pi(p-1)^2+(1-\pi)p^2,
\]
with derivative $\partial_p \mathbb{E}[L]=2(p-\pi)$ and positive second derivative. Thus, expected loss is uniquely minimized at $p=\pi$, so a solver optimally reports its true probability that $y=1$.

Recall that the scoring target throughout the experiments is the proposer's private commitment $y^{(r)}_i$. Even in verifiable domains, such as Python code challenges where execution could produce the correct answer, the solvers are scored against $y^{(r)}_i$.
The roles of proposer and solver therefore ask different things of a model, namely to find questions that separate the population and to forecast other proposers' committed answers.

Proposer and solver scores accumulate as 2 separate averages. The proposer score of model $i$
up to round $r$, \(\bar P^{(r)}_i\), is the mean of its scored proposal rewards.
Its solver loss, \(\bar L^{(r)}_i\), is the mean of its scored response losses.
Clearly, higher \(\bar P^{(r)}_i\) and lower \(\bar L^{(r)}_i\) are better. Let $P_i = \bar P^{(R)}_i$ and $L_i = \bar L^{(R)}_i$ be the final scores.
For the final ranking, we combine the 2 components with equal weight into the joint score
\[
J_i=\frac{ P_i+(1- L_i)}{2}.
\]
In Section~\ref{sec:experiments} we report both $J_i$ and the separate component scores $P_i, \, L_i$ for each model $i$, separately for the open-ended and code-constrained SepaRank games.

\subsection{Adaptive solver weights}
\label{subsec:adaptive-weighting}

Assume the solvers in $[N] \setminus \{i\}$ can be separated into 2 groups $A$ and $B$,
each consisting of solvers of similar ability, with those in $A$ significantly more capable than those in $B$, and proposer $i$ has identified this separation.
With equal weights $w^{(r)}_{-i}$, proposer $i$ can keep earning a high reward by
separating groups $A$ and $B$. Once a question distinguishes
the stronger solvers from the rest, variations of that question may
keep producing disagreement without revealing finer differences among the
stronger solvers.

We introduce adaptive weights that give more importance, in computing the proposer's score, to models that have performed well as
solvers.
Note that weights determine solely how much each solver's report contributes to
the proposer's reward, and do not affect the solver's incurred loss.
Hence, a model with a small weight retains the same opportunities
to improve its solver score.

In practice, SepaRank begins with 3 warmup rounds in which all \(N\) models
have equal weights. These initial interactions aim to make past performance sufficiently informative of the models' actual capabilities before it can influence rewards.
That is, \(w^{(r)}_i = 1/N\) for \(r \leq 3\). After each round \(r \geq 3\), we use model $i$'s cumulative mean solver loss \(\bar L^{(r)}_i\)
to set its weight for the next round as
\[
w^{(r+1)}_i
=
0.95\,
\frac{\exp(-\bar L^{(r)}_i/0.25)}
     {\sum_{k=1}^{N}\exp(-\bar L^{(r)}_k/0.25)}
+
\frac{0.05}{N}.
\]
Note that lower solver loss gives greater weight. The temperature parameter \(0.25\) is set empirically and
controls how sharply the weights respond to differences in solver scores.
A smaller value would place more weight on the best-performing
solvers, whereas a larger one would distribute weight more evenly. 
We distribute \(5\%\) of the weight uniformly among solvers, to ensure
that every model retains some influence on the reward.

The intended effect of adaptive weights is to make broad, established separations
less rewarding in future rounds.
If most of the weight lies on solvers that answer similarly and the
disagreement is confined to low-weight solvers, the proposer reward is small. We conjecture that this encourages finer
separation among similarly capable solvers. We do not claim that this rule explicitly
rewards novelty, as this would require a formal definition of ``novelty'' and a judge to enforce it.

\subsection{Resolution rules and optional adjudication}
\label{subsec:resolution-rules}
We have introduced SepaRank as a judge-free game, in which the resolution of a proposed question is the responsibility of the proposer. By allowing questions to be open-ended and potentially unverifiable, we preclude the use of a formal, external resolution rule which would supply the ``correct'' answer $y_i$ to any posed question $q_i$. Nevertheless, one might wish to constrain the game to verifiable domains, in which such a rule can be formally defined and is computable. In those cases, \(y\) may be determined by executing a program, checking a proof, 
or verifying a supplied certificate. Here, the solver loss would measure the accuracy of the solver with respect to an objective public truth.
In non-verifiable domains, where \(y\) may
instead be the proposer's committed answer, one could ask the proposer to supply a rationale for their private commitment, to be checked sparsely by a judge. We do not pursue this extension in SepaRank.

If the proposer is allowed to commit any $y$ they wish, the solver must reason not only about a potential objectively correct answer but also about the possibility that the proposer might, either accidentally or deliberately, commit an erroneous answer. The incentives this creates for the proposer, and the extent to which models exploit them, are measured in Section~\ref{subsec:honesty}. In experiments where we restrict the question scope to Python programs, we can verify the true $y$ by execution, and use it to understand the commitment behavior of proposers.

Although false commitments may increase disagreement in the short run, we conjecture that they should eventually erode trust.\footnote{Note the proposer's reward $P_i^{(r)}$ does not depend on the committed answer $y^{(r)}_i$. Therefore, misrepresenting $y^{(r)}_i$ is a strategic choice that factors in how solvers will respond in future rounds to the miscommitment at round $r$.} As a proposer becomes less reliable, calibrated solvers ought to hedge toward $\frac{1}{2}$, reducing the proposer's variance reward and thus creating a long-run incentive for honest commitments. The theoretical foundations of reputation, strategic stability, and conditions favoring semantically faithful commitments will be explored in future work.

Altogether, the proposer score always has the same meaning. It measures the ability to
find questions on which the solver population separates. The meaning of the solver score,
however, depends on the chosen resolution rule.

\raggedbottom
\setlength{\intextsep}{5pt plus 1pt minus 1pt}
\setlength{\textfloatsep}{7pt plus 1pt minus 1pt}
\setlength{\floatsep}{6pt plus 1pt minus 1pt}
\setlength{\abovecaptionskip}{4pt}
\renewcommand{\topfraction}{.95}
\renewcommand{\bottomfraction}{.95}
\renewcommand{\textfraction}{.05}
\renewcommand{\floatpagefraction}{.80}
\setcounter{topnumber}{6}\setcounter{bottomnumber}{6}\setcounter{totalnumber}{10}
\makeatletter
\setlength{\@fptop}{0pt}
\setlength{\@fpsep}{7pt}
\setlength{\@fpbot}{0pt plus 1fil}
\makeatother

\definecolor{srCode}{HTML}{176A93}
\definecolor{srOpen}{HTML}{C05B21}
\definecolor{srJoint}{HTML}{75488F}
\pgfplotsset{
  sr/base/.style={scale only axis,axis lines=left,
    axis line style={black!35},tick style={black!35},
    tick label style={font=\scriptsize},label style={font=\footnotesize},
    title style={font=\footnotesize\bfseries},
    grid=major,grid style={black!8},scaled ticks=false,
    tick align=outside,clip=false},
  sr/forest/.style={sr/base,y dir=reverse,ymajorgrids=false,
    ytick style={draw=none},axis y line=left,y axis line style={draw=none}},
}
\definecolor{srModel0}{HTML}{332288}
\definecolor{srModel1}{HTML}{0077AA}
\definecolor{srModel2}{HTML}{117733}
\definecolor{srModel3}{HTML}{44AA99}
\definecolor{srModel4}{HTML}{997700}
\definecolor{srModel5}{HTML}{CC6677}
\definecolor{srModel6}{HTML}{882255}
\definecolor{srModel7}{HTML}{AA4499}
\definecolor{srModel8}{HTML}{6699CC}
\definecolor{srModel9}{HTML}{555555}

\section{Experiments}
\label{sec:experiments}

We run SepaRank in 2 domains. In \emph{code-constrained games}, proposers submit a Python program with a no-argument \texttt{main()} returning $0$ or $1$. In \emph{open-ended games}, they submit a yes/no question. In both, solvers predict the proposer's private commitment. The code execution checks admissibility, but its output does not replace the proposer's commitment $y_i$, used to score the solvers. 

We run 10 independent games per domain, each lasting 20 rounds and containing the same 10 models from 5 providers (Table~\ref{tab:roster}). Every model proposes once per round, and all 9 peers solve each valid challenge. Players have stable pseudonyms within a game, and the current proposer's pseudonym is disclosed during a match. Prompts give the players the full protocol details (scoring rules, constraints, cumulative standings, etc.) and the history of the last 5 completed rounds, which includes proposed questions with their proposers and commitments, and players' answers. The history is truncated to 5 rounds to reduce API expenses. We present the full setup and the prompts in Appendices~\ref{a-implementation} and~\ref{a-prompts}, respectively.

For each model $i$ we report the final proposer score $P_i=\bar P^{(R)}_i$ and solver loss $L_i=\bar L^{(R)}_i$ of Section~\ref{sec:protocol}, with $R=20$, averaged over the games of each domain. We also define the solver skill $S_i=1-L_i$, so that the joint score is $J_i=(P_i+S_i)/2$. We also present error intervals, which are pointwise $95\%$ percentile intervals from $50{,}000$ whole-game bootstrap resamples.

In Appendix~\ref{a-additional_experiments} we explore the score dynamics
of games, the false commitment behavior of proposers in the verifiable
code-constrained setting, and the anonymous author ablation, which hides the
proposer's name in a match.

\subsection{Rankings}
\label{subsec:rankings-committed}

We report separate joint-score rankings for both the code-constrained and open-ended domains in Figure~\ref{fig:joint-rankings}. We break down the components in more detail in Table~\ref{tab:full-results}. 

For both rankings, GPT-6 Astra ranks 1st with $J=0.762$ for code-constrained and $0.732$ for open-ended games, and Claude Fable 5.1 ranks 2nd, with $0.640$ and $0.667$. 

There are some small differences in the rest of the model rankings, but the results are generally consistent with what we expect in terms of model capabilities. 

\begin{figure}[!htbp]
\centering
\begin{tikzpicture}
\begin{axis}[sr/forest,width=86pt,height=116pt,
at={(0pt,0)},anchor=north west,
xmin=.34,xmax=.82,ymin=-.65,ymax=9.65,xtick={.4,.6,.8},ytick={0,...,9},
yticklabels={{GPT-6 Astra},{Claude Fable 5.1},{GPT-5.6 Terra},{Gemini 3.8 Flash},{Grok 4.6},{Grok 4.5},{Muse Spark 1.2},{Muse Spark 1.1},{Claude Sonnet 5},{Gemini 3.1 Pro Preview}},
yticklabel style={font=\scriptsize},title={Code-constrained},xlabel={Joint score $J$}]
\path[fill=black!3] (axis cs:.34,-0.47) rectangle (axis cs:.82,0.47);
\path[fill=black!3] (axis cs:.34,1.53) rectangle (axis cs:.82,2.4699999999999998);
\path[fill=black!3] (axis cs:.34,3.5300000000000002) rectangle (axis cs:.82,4.47);
\path[fill=black!3] (axis cs:.34,5.53) rectangle (axis cs:.82,6.47);
\path[fill=black!3] (axis cs:.34,7.53) rectangle (axis cs:.82,8.47);
\addplot[only marks,mark=*,mark size=2pt,color=srCode,error bars/.cd,x dir=both,x explicit,error bar style={line width=.75pt},error mark options={rotate=90,mark size=1.7pt}]
table[x=x,y=y,x error minus=em,x error plus=ep] {
x y em ep
0.7617912894 0 0.0200283082 0.0208584222
0.6399003115 1 0.0282021938 0.0292099417
0.5584729759 2 0.0290763455 0.0285343507
0.5467029773 3 0.0190611418 0.0191888418
0.5129035565 4 0.0192022038 0.0206639584
0.5026134488 5 0.0570324843 0.0570503347
0.4962044268 6 0.0222686829 0.0191157565
0.4839045190 7 0.0234724926 0.0258778997
0.4012621254 8 0.0258355007 0.0283252366
0.3788454019 9 0.0156734466 0.0161146231
};
\node[anchor=west,font=\scriptsize,text=srCode] at (axis cs:.827,0) {0.762};
\node[anchor=west,font=\scriptsize,text=srCode] at (axis cs:.827,1) {0.640};
\node[anchor=west,font=\scriptsize,text=srCode] at (axis cs:.827,2) {0.558};
\node[anchor=west,font=\scriptsize,text=srCode] at (axis cs:.827,3) {0.547};
\node[anchor=west,font=\scriptsize,text=srCode] at (axis cs:.827,4) {0.513};
\node[anchor=west,font=\scriptsize,text=srCode] at (axis cs:.827,5) {0.503};
\node[anchor=west,font=\scriptsize,text=srCode] at (axis cs:.827,6) {0.496};
\node[anchor=west,font=\scriptsize,text=srCode] at (axis cs:.827,7) {0.484};
\node[anchor=west,font=\scriptsize,text=srCode] at (axis cs:.827,8) {0.401};
\node[anchor=west,font=\scriptsize,text=srCode] at (axis cs:.827,9) {0.379};
\end{axis}
\begin{axis}[sr/forest,width=86pt,height=116pt,
at={(196pt,0)},anchor=north west,
xmin=.34,xmax=.82,ymin=-.65,ymax=9.65,xtick={.4,.6,.8},ytick={0,...,9},
yticklabels={{GPT-6 Astra},{Claude Fable 5.1},{Gemini 3.8 Flash},{GPT-5.6 Terra},{Muse Spark 1.2},{Grok 4.6},{Muse Spark 1.1},{Grok 4.5},{Gemini 3.1 Pro Preview},{Claude Sonnet 5}},
yticklabel style={font=\scriptsize},title={Open-ended},xlabel={Joint score $J$}]
\path[fill=black!3] (axis cs:.34,-0.47) rectangle (axis cs:.82,0.47);
\path[fill=black!3] (axis cs:.34,1.53) rectangle (axis cs:.82,2.4699999999999998);
\path[fill=black!3] (axis cs:.34,3.5300000000000002) rectangle (axis cs:.82,4.47);
\path[fill=black!3] (axis cs:.34,5.53) rectangle (axis cs:.82,6.47);
\path[fill=black!3] (axis cs:.34,7.53) rectangle (axis cs:.82,8.47);
\addplot[only marks,mark=*,mark size=2pt,color=srOpen,error bars/.cd,x dir=both,x explicit,error bar style={line width=.75pt},error mark options={rotate=90,mark size=1.7pt}]
table[x=x,y=y,x error minus=em,x error plus=ep] {
x y em ep
0.7317306336 0 0.0232379868 0.0203023305
0.6665340902 1 0.0331950181 0.0277440927
0.5652999616 2 0.0288494733 0.0255536328
0.5569274265 3 0.0224385418 0.0232029001
0.5408563945 4 0.0227598413 0.0260954669
0.5250218176 5 0.0200242910 0.0210985057
0.5222490690 6 0.0168894285 0.0178446410
0.5072952820 7 0.0127142066 0.0129342796
0.4614538056 8 0.0203424657 0.0240193404
0.3954355808 9 0.0184555787 0.0197147539
};
\node[anchor=west,font=\scriptsize,text=srOpen] at (axis cs:.827,0) {0.732};
\node[anchor=west,font=\scriptsize,text=srOpen] at (axis cs:.827,1) {0.667};
\node[anchor=west,font=\scriptsize,text=srOpen] at (axis cs:.827,2) {0.565};
\node[anchor=west,font=\scriptsize,text=srOpen] at (axis cs:.827,3) {0.557};
\node[anchor=west,font=\scriptsize,text=srOpen] at (axis cs:.827,4) {0.541};
\node[anchor=west,font=\scriptsize,text=srOpen] at (axis cs:.827,5) {0.525};
\node[anchor=west,font=\scriptsize,text=srOpen] at (axis cs:.827,6) {0.522};
\node[anchor=west,font=\scriptsize,text=srOpen] at (axis cs:.827,7) {0.507};
\node[anchor=west,font=\scriptsize,text=srOpen] at (axis cs:.827,8) {0.461};
\node[anchor=west,font=\scriptsize,text=srOpen] at (axis cs:.827,9) {0.395};
\end{axis}
\end{tikzpicture}
\caption{Model rankings according to joint score $J$ over 10 games of 20 rounds per domain. Error bars are $95\%$ game-bootstrap intervals.}
\label{fig:joint-rankings}
\end{figure}

\begin{table}[!htbp]
\centering
\caption{Proposer, solver, and joint scores. Rows follow the code-constrained rank, and the last column gives the open-ended rank.}
\label{tab:full-results}\label{tab:roster}
\footnotesize\setlength{\tabcolsep}{2.2pt}\renewcommand{\arraystretch}{1.07}
\begin{tabular}{rllrrrrrrr}\toprule
& & & \multicolumn{3}{c}{Code-constrained} & \multicolumn{3}{c}{Open-ended} & \\
\cmidrule(lr){4-6}\cmidrule(lr){7-9}
Rank & Model & Provider & $P\uparrow$ & $L\downarrow$ & $J\uparrow$ & $P\uparrow$ & $L\downarrow$ & $J\uparrow$ & Rank \\
\midrule
1 & GPT-6 Astra & OpenAI & 0.629 & 0.105 & 0.762 & 0.568 & 0.104 & 0.732 & 1 \\
2 & Claude Fable 5.1 & Anthropic & 0.401 & 0.121 & 0.640 & 0.424 & 0.091 & 0.667 & 2 \\
3 & GPT-5.6 Terra & OpenAI & 0.364 & 0.247 & 0.558 & 0.316 & 0.202 & 0.557 & 4 \\
4 & Gemini 3.8 Flash & Google & 0.274 & 0.181 & 0.547 & 0.262 & 0.131 & 0.565 & 3 \\
5 & Grok 4.6 & xAI & 0.267 & 0.241 & 0.513 & 0.210 & 0.160 & 0.525 & 6 \\
6 & Grok 4.5 & xAI & 0.254 & 0.249 & 0.503 & 0.183 & 0.169 & 0.507 & 8 \\
7 & Muse Spark 1.2 & Meta & 0.236 & 0.244 & 0.496 & 0.244 & 0.162 & 0.541 & 5 \\
8 & Muse Spark 1.1 & Meta & 0.238 & 0.270 & 0.484 & 0.241 & 0.197 & 0.522 & 7 \\
9 & Claude Sonnet 5 & Anthropic & 0.142 & 0.339 & 0.401 & 0.054 & 0.263 & 0.395 & 10 \\
10 & Gemini 3.1 Pro Preview & Google & 0.057 & 0.299 & 0.379 & 0.096 & 0.173 & 0.461 & 9 \\
\bottomrule\end{tabular}
\end{table}

The role profiles (Figure~\ref{fig:role-profiles}) break the joint score down further into proposer and solver scores. GPT-6 Astra has the highest proposer reward in both the code-constrained and open-ended domains. It is also the best solver in the code-constrained domain, while Claude Fable 5.1 is the best solver in the open-ended domain.
GPT-5.6 Terra is a better proposer than Gemini 3.8 Flash in both domains, but Gemini 3.8 Flash is the better solver. 
In general, we observe that the proposer and solver scores contribute distinct information about a model's capabilities.

\begin{figure}[!htbp]
\centering
\begin{tikzpicture}
\begin{axis}[sr/forest,width=107pt,height=118pt,at={(0pt,0)},anchor=north west,
xmin=0,xmax=0.72,xtick={0,.2,.4,.6},ymin=-.6,ymax=9.6,ytick={0,...,9},
yticklabels={{GPT-6 Astra},{GPT-5.6 Terra},{Claude Fable 5.1},{Claude Sonnet 5},{Gemini 3.1 Pro Preview},{Gemini 3.8 Flash},{Grok 4.6},{Grok 4.5},{Muse Spark 1.1},{Muse Spark 1.2}},
title={Proposer reward $P$}]
\path[fill=black!3] (axis cs:0,-0.45) rectangle (axis cs:0.72,0.45);
\path[fill=black!3] (axis cs:0,1.55) rectangle (axis cs:0.72,2.45);
\path[fill=black!3] (axis cs:0,3.55) rectangle (axis cs:0.72,4.45);
\path[fill=black!3] (axis cs:0,5.55) rectangle (axis cs:0.72,6.45);
\path[fill=black!3] (axis cs:0,7.55) rectangle (axis cs:0.72,8.45);
\addplot[only marks,mark=*,mark size=2pt,color=srCode,error bars/.cd,x dir=both,x explicit,error bar style={line width=.75pt},error mark options={rotate=90,mark size=1.7pt}]
table[x=x,y=y,x error minus=em,x error plus=ep] {
x y em ep
0.6285700459 -0.1300000000 0.0333151557 0.0329298232
0.3637415741 0.8700000000 0.0675545906 0.0686510081
0.4010737793 1.8700000000 0.0601359927 0.0598133098
0.1415014331 2.8700000000 0.0270488325 0.0245613850
0.0566507978 3.8700000000 0.0149665980 0.0139396270
0.2739101481 4.8700000000 0.0455616763 0.0389180894
0.2668482558 5.8700000000 0.0389155412 0.0423490562
0.2544931133 6.8700000000 0.1074424282 0.1107447470
0.2375652576 7.8700000000 0.0615374558 0.0670166323
0.2362253706 8.8700000000 0.0487747011 0.0397770179
};
\addplot[only marks,mark=diamond*,mark size=2pt,color=srOpen,error bars/.cd,x dir=both,x explicit,error bar style={line width=.75pt},error mark options={rotate=90,mark size=1.7pt}]
table[x=x,y=y,x error minus=em,x error plus=ep] {
x y em ep
0.5675504649 0.1300000000 0.0679323303 0.0548025573
0.3160527447 1.1300000000 0.0688101748 0.0689637741
0.4239321425 2.1300000000 0.0803940990 0.0674610632
0.0541030115 3.1300000000 0.0309999237 0.0368856344
0.0962268773 4.1300000000 0.0509264697 0.0760236657
0.2615579693 5.1300000000 0.0701377556 0.0599468645
0.2103519379 6.1300000000 0.0646444699 0.0624597847
0.1833963756 7.1300000000 0.0589233609 0.0630028985
0.2413426598 8.1300000000 0.0658079428 0.0648830154
0.2437459283 9.1300000000 0.0622268880 0.0693274519
};
\end{axis}
\begin{axis}[sr/forest,width=107pt,height=118pt,at={(126pt,0)},anchor=north west,
xmin=0.6,xmax=0.97,xtick={.6,.7,.8,.9},ymin=-.6,ymax=9.6,ytick={0,...,9},
yticklabels={{},{},{},{},{},{},{},{},{},{}},
title={Solver skill $S=1-L$}]
\path[fill=black!3] (axis cs:0.6,-0.45) rectangle (axis cs:0.97,0.45);
\path[fill=black!3] (axis cs:0.6,1.55) rectangle (axis cs:0.97,2.45);
\path[fill=black!3] (axis cs:0.6,3.55) rectangle (axis cs:0.97,4.45);
\path[fill=black!3] (axis cs:0.6,5.55) rectangle (axis cs:0.97,6.45);
\path[fill=black!3] (axis cs:0.6,7.55) rectangle (axis cs:0.97,8.45);
\addplot[only marks,mark=*,mark size=2pt,color=srCode,error bars/.cd,x dir=both,x explicit,error bar style={line width=.75pt},error mark options={rotate=90,mark size=1.7pt}]
table[x=x,y=y,x error minus=em,x error plus=ep] {
x y em ep
0.8950125329 -0.1300000000 0.0201984335 0.0190112080
0.7532043777 0.8700000000 0.0252581700 0.0238284060
0.8787268437 1.8700000000 0.0183507036 0.0178607381
0.6610228178 2.8700000000 0.0332592364 0.0370087681
0.7010400060 3.8700000000 0.0243201725 0.0241927974
0.8194958064 4.8700000000 0.0275325337 0.0289288119
0.7589588572 5.8700000000 0.0282408393 0.0270766070
0.7507337843 6.8700000000 0.0181154549 0.0164279668
0.7302437804 7.8700000000 0.0248310229 0.0248730244
0.7561834830 8.8700000000 0.0195025143 0.0197958472
};
\addplot[only marks,mark=diamond*,mark size=2pt,color=srOpen,error bars/.cd,x dir=both,x explicit,error bar style={line width=.75pt},error mark options={rotate=90,mark size=1.7pt}]
table[x=x,y=y,x error minus=em,x error plus=ep] {
x y em ep
0.8959108022 0.1300000000 0.0387233033 0.0328333372
0.7978021083 1.1300000000 0.0455926777 0.0411318983
0.9091360378 2.1300000000 0.0303148528 0.0247907806
0.7367681500 3.1300000000 0.0423992506 0.0449844749
0.8266807339 4.1300000000 0.0540225526 0.0446689248
0.8690419539 5.1300000000 0.0395915097 0.0331115483
0.8396916972 6.1300000000 0.0425178367 0.0360360356
0.8311941883 7.1300000000 0.0435831592 0.0394472114
0.8031554783 8.1300000000 0.0516979861 0.0478467398
0.8379668606 9.1300000000 0.0395177294 0.0351153639
};
\end{axis}
\end{tikzpicture}
\par\smallskip{\footnotesize\textcolor{srCode}{$\bullet$ Code-constrained}\qquad\textcolor{srOpen}{$\blacklozenge$ Open-ended}}
\caption{Proposer and solver scores across code-constrained and open-ended domains, with error bars showing $95\%$ game-bootstrap intervals.}
\label{fig:role-profiles}
\end{figure}

The cross-domain plot (Figure~\ref{fig:pcqc}) shows broad agreement, with Spearman $\rho=0.915$ between the 10 models' joint scores. Component correlations are $0.879$ for proposer reward and $0.903$ for solver skill. This agreement coexists with domain-specific differences. For example, Muse Spark 1.2 ranks 7th in code-constrained and 5th in open-ended games.

\begin{figure}[!htbp]
\centering
\begingroup
\definecolor{srCrossPoint}{HTML}{176A93}
\definecolor{srCrossInterval}{HTML}{7CABC0}
\begin{tikzpicture}
\begin{axis}[
 width=.76\linewidth,height=112pt,scale only axis,
 xmin=.35,xmax=.82,ymin=.34,ymax=.80,
 xlabel={Code-constrained joint score $J$},ylabel={Open-ended joint score $J$},
 xtick={.4,.5,.6,.7,.8},ytick={.4,.5,.6,.7,.8},
 scaled ticks=false,tick label style={font=\scriptsize},
 label style={font=\small},title style={font=\small},
 title={Agreement across domains ($\rho=0.915$)},
 axis lines=left,axis line style={black!40},tick style={black!40},
 grid=major,grid style={black!8},clip=false,
 every axis plot/.append style={line width=.6pt},
]
\addplot[gray!55,densely dashed,no marks] coordinates {(.35,.35) (.80,.80)};
\addplot[only marks,mark=*,mark size=2pt,srCrossPoint,
 error bars/.cd,x dir=both,x explicit,y dir=both,y explicit,
 error bar style={srCrossInterval,line width=.55pt},error mark options={rotate=90,mark size=1.8pt}]
 table[x=x,y=y,x error minus=xminus,x error plus=xplus,
 y error minus=yminus,y error plus=yplus,row sep=\\] {
x y xminus xplus yminus yplus\\
0.761791289418 0.731730633567 0.020028308189 0.020858422178 0.023237986808 0.020302330460\\ 
0.558472975900 0.556927426526 0.029076345549 0.028534350736 0.022438541779 0.023202900122\\ 
0.639900311503 0.666534090156 0.028202193821 0.029209941735 0.033195018115 0.027744092685\\ 
0.401262125427 0.395435580759 0.025835500667 0.028325236583 0.018455578698 0.019714753923\\ 
0.378845401903 0.461453805603 0.015673446561 0.016114623122 0.020342465738 0.024019340361\\ 
0.546702977263 0.565299961595 0.019061141763 0.019188841755 0.028849473333 0.025553632784\\ 
0.512903556507 0.525021817583 0.019202203775 0.020663958439 0.020024291010 0.021098505666\\ 
0.502613448763 0.507295281980 0.057032484312 0.057050334689 0.012714206612 0.012934279610\\ 
0.483904519010 0.522249069043 0.023472492616 0.025877899747 0.016889428497 0.017844640964\\ 
0.496204426833 0.540856394452 0.022268682857 0.019115756468 0.022759841323 0.026095466930\\ 
};
\node[anchor=west,font=\scriptsize,text=black!85,fill=white,inner sep=1.3pt] (srCrossLabel0) at (axis cs:0.684,0.776) {GPT-6 Astra};
\draw[black!40,line width=.35pt] (axis cs:0.761791289418,0.731730633567) -- (srCrossLabel0);
\node[anchor=west,font=\scriptsize,text=black!85,fill=white,inner sep=1.3pt] (srCrossLabel1) at (axis cs:0.64,0.583) {GPT-5.6 Terra};
\draw[black!40,line width=.35pt] (axis cs:0.558472975900,0.556927426526) -- (srCrossLabel1);
\node[anchor=west,font=\scriptsize,text=black!85,fill=white,inner sep=1.3pt] (srCrossLabel2) at (axis cs:0.58,0.714) {Claude Fable 5.1};
\draw[black!40,line width=.35pt] (axis cs:0.639900311503,0.666534090156) -- (srCrossLabel2);
\node[anchor=west,font=\scriptsize,text=black!85,fill=white,inner sep=1.3pt] (srCrossLabel3) at (axis cs:0.43,0.361) {Claude Sonnet 5};
\draw[black!40,line width=.35pt] (axis cs:0.401262125427,0.395435580759) -- (srCrossLabel3);
\node[anchor=west,font=\scriptsize,text=black!85,fill=white,inner sep=1.3pt] (srCrossLabel4) at (axis cs:0.365,0.506) {Gemini 3.1 Pro Preview};
\draw[black!40,line width=.35pt] (axis cs:0.378845401903,0.461453805603) -- (srCrossLabel4);
\node[anchor=west,font=\scriptsize,text=black!85,fill=white,inner sep=1.3pt] (srCrossLabel5) at (axis cs:0.36,0.682) {Gemini 3.8 Flash};
\draw[black!40,line width=.35pt] (axis cs:0.546702977263,0.565299961595) -- (srCrossLabel5);
\node[anchor=west,font=\scriptsize,text=black!85,fill=white,inner sep=1.3pt] (srCrossLabel6) at (axis cs:0.64,0.525) {Grok 4.6};
\draw[black!40,line width=.35pt] (axis cs:0.512903556507,0.525021817583) -- (srCrossLabel6);
\node[anchor=west,font=\scriptsize,text=black!85,fill=white,inner sep=1.3pt] (srCrossLabel7) at (axis cs:0.63,0.464) {Grok 4.5};
\draw[black!40,line width=.35pt] (axis cs:0.502613448763,0.507295281980) -- (srCrossLabel7);
\node[anchor=west,font=\scriptsize,text=black!85,fill=white,inner sep=1.3pt] (srCrossLabel8) at (axis cs:0.36,0.57) {Muse Spark 1.1};
\draw[black!40,line width=.35pt] (axis cs:0.483904519010,0.522249069043) -- (srCrossLabel8);
\node[anchor=west,font=\scriptsize,text=black!85,fill=white,inner sep=1.3pt] (srCrossLabel9) at (axis cs:0.36,0.622) {Muse Spark 1.2};
\draw[black!40,line width=.35pt] (axis cs:0.496204426833,0.540856394452) -- (srCrossLabel9);
\end{axis}
\end{tikzpicture}
\endgroup
\caption{Open-ended versus code-constrained joint score. The error bars are $95\%$ game-bootstrap intervals on both axes. The rank correlation is $\rho=0.915$.}
\label{fig:pcqc}
\end{figure}

\subsection{Comparison with external benchmarks}
\label{subsec:external-benchmarks}

We compare SepaRank with 10 individual benchmarks and the Artificial Analysis
Intelligence Index for the
same 10 model releases \citep{review_artificialanalysis2026}.
The 10 benchmarks are the components of the Artificial Analysis Intelligence Index v4.3.2, taken without selection. It is important to note that the benchmarks all measure different aspects of model capability. For example, AA-Omniscience penalizes confident wrong answers and so partly rewards abstention in addition to knowledge alone.

Figure~\ref{fig:benchmark-correlations} reports Spearman correlations
separately for proposer reward, solver skill, and joint score.
The Index correlations are $0.900$, $0.900$, and $0.936$ for code-constrained games,
and $0.790$, $0.717$, and $0.851$ for open-ended games.

The correlation with SepaRank varies across benchmarks. For example, the code-constrained joint score correlates
$0.915$ with Terminal-Bench 4.0, but $-0.006$ with SciCode.

\begin{figure}[!htbp]
\centering
\begingroup
\definecolor{SRcorrBlue}{HTML}{2166AC}
\definecolor{SRcorrRed}{HTML}{B2182B}

\newcommand{\SRcorrRow}[3]{%
  \pgfmathsetmacro{\SRcorrY}{144-12*(#1)-4*(#1>=10)}%
  \node[anchor=east,inner sep=0pt,font=\fontsize{8}{9}\selectfont]
    at (-8,\SRcorrY) {#2};
  \foreach \SRcorrValue [count=\SRcorrColumn from 0] in {#3}{%
    \pgfmathsetmacro{\SRcorrX}{41*\SRcorrColumn+7*(\SRcorrColumn>2)}%
    \pgfmathsetmacro{\SRcorrMix}{100*abs(\SRcorrValue)}%
    \ifdim\SRcorrValue pt<0pt
      \def\SRcorrHue{SRcorrRed}%
    \else
      \def\SRcorrHue{SRcorrBlue}%
    \fi
    \fill[\SRcorrHue!\SRcorrMix!white]
      (\SRcorrX,\SRcorrY-5.85) rectangle +(41,11.7);
    \ifdim\SRcorrMix pt>55pt
      \def\SRcorrInk{white}%
    \else
      \def\SRcorrInk{black}%
    \fi
    \def\SRcorrWeight{}%
    \ifnum\SRcorrColumn=2
      \def\SRcorrWeight{\bfseries\boldmath}%
    \fi
    \ifnum\SRcorrColumn=5
      \def\SRcorrWeight{\bfseries\boldmath}%
    \fi
    \node[text=\SRcorrInk,inner sep=0pt,
      font=\fontsize{8}{9}\selectfont\SRcorrWeight]
      at (\SRcorrX+20.5,\SRcorrY)
      {$\pgfmathprintnumber[fixed,precision=2,zerofill]{\SRcorrValue}$};
  }%
}

\begin{tikzpicture}[x=.96pt,y=.93pt]
  \useasboundingbox (-143,-15) rectangle (254,180);

  \node[font=\small\bfseries,inner sep=0pt]
    at (61.5,174) {Code-constrained};
  \node[font=\small\bfseries,inner sep=0pt]
    at (191.5,174) {Open-ended};

  \foreach \x/\score in
    {20.5/P,61.5/S,102.5/J,150.5/P,191.5/S,232.5/J}
    \node[font=\small,inner sep=0pt] at (\x,159) {$\score$};

  \SRcorrRow{0}{AA-Briefcase v1.1}
    {0.575757576,0.600000000,0.624242424,0.454545455,0.503030303,0.503030303}
  \SRcorrRow{1}{GDPval-AA v2.1}
    {0.563636364,0.672727273,0.624242424,0.490909091,0.575757576,0.563636364}
  \SRcorrRow{2}{AutomationBench-AA}
    {0.875383983,0.838909650,0.887542094,0.705170431,0.601826488,0.778119096}
  \SRcorrRow{3}{Terminal-Bench 4.0}
    {0.903030303,0.757575758,0.915151515,0.721212121,0.527272727,0.757575758}
  \SRcorrRow{4}{SciCode}
    {0.018406294,0.257688110,-0.006135431,0.282229835,0.472428202,0.282229835}
  \SRcorrRow{5}{Humanity's Last Exam}
    {0.472428202,0.644220275,0.460157339,0.656491137,0.754658037,0.681032862}
  \SRcorrRow{6}{GDP.pdf}
    {0.826751539,0.741644763,0.844988706,0.796356262,0.632221765,0.802435318}
  \SRcorrRow{7}{CritPt}
    {0.613147101,0.631727316,0.668887746,0.761788822,0.464505379,0.755595417}
  \SRcorrRow{8}{AA-Omniscience}
    {0.359762786,0.603669759,0.347567437,0.408544180,0.798795338,0.493911621}
  \SRcorrRow{9}{AA-LCR v1.1}
    {0.287473872,0.122329307,0.305823268,0.232425684,0.061164654,0.226309218}

  \draw[black!25,line width=.3pt] (-136,28)--(253,28);
  \SRcorrRow{10}{AA Intelligence Index}
    {0.899700205,0.899700205,0.936174537,0.790277207,0.717328542,0.851067761}

  \shade[left color=SRcorrRed,right color=white]
    (0,3) rectangle (126.5,8);
  \shade[left color=white,right color=SRcorrBlue]
    (126.5,3) rectangle (253,8);
  \foreach \x/\v in {0/-1,63.25/-0.5,126.5/0,189.75/0.5,253/1}
    \node[anchor=north,inner sep=1pt,font=\fontsize{7}{8}\selectfont]
      at (\x,2) {$\v$};
  \node[font=\fontsize{8}{9}\selectfont,inner sep=0pt]
    at (126.5,-11) {Spearman correlation $\rho$};
\end{tikzpicture}
\endgroup

\caption{Spearman correlations across the 10 models for proposer
reward $P$, solver skill $S=1-L$, and joint score $J$.
The bottom row is the Artificial Analysis
Intelligence Index, an aggregate of the 10 individual benchmarks.
Data from Artificial Analysis v4.3.2, accessed 23 September 2026.}
\label{fig:benchmark-correlations}
\end{figure}

SepaRank scores correlate positively with every individual benchmark except SciCode in the code-constrained joint score. The composite Artificial Analysis (AA) Intelligence Index gives the highest correlation with the joint score in both domains ($0.936$ for code-constrained games, $0.851$ for open-ended games). 
The correlations obtained from experiments with these 10 models indicate that SepaRank, a protocol that requires neither human-authored questions nor a semantic judge, can closely track existing measures of capability.

Correlations with the Index cannot be compared fairly across items, since each benchmark is itself part of the Index. Instead, we estimate the $g$-loading $\ell_k$ of each item (Section~\ref{sec:spearman}) with a leave-one-out proxy for $g$. The benchmarks show a positive manifold, with 41 of their 45 pairwise Spearman correlations positive, which motivates an equal-weight composite as the proxy. We treat the SepaRank joint score $J$ as an 11th item alongside the 10 AA benchmarks. For each item $k$, we standardize each of the other 10 items across the 10 models and average them with equal weights into a proxy $\hat g_{-k}$, and we estimate $\ell_k$ by the Spearman correlation between item $k$ and $\hat g_{-k}$. The SepaRank joint score has an estimated $g$-loading at least as high as that of the best individual benchmark (Table~\ref{tab:item-rest}).

\begin{table}[!htbp]
\centering\small
\caption{Estimated $g$-loadings, given by the Spearman correlation across the 10 models between each item $k$ and its leave-one-out proxy $\hat g_{-k}$, the equal-weight average of the other 10 items. The proxy averages the other items either as standardized scores ($z$) or as ranks.}
\label{tab:item-rest}
\begin{tabular}{@{}lrrrr@{}}
\toprule
& \multicolumn{2}{c}{Code-constrained} & \multicolumn{2}{c}{Open-ended} \\
\cmidrule(lr){2-3}\cmidrule(lr){4-5}
Item & $z$ & Rank & $z$ & Rank \\
\midrule
SepaRank $J$ & $\mathbf{0.94}$ & $\mathbf{0.81}$ & $\mathbf{0.88}$ & $\mathbf{0.84}$ \\
\midrule
Terminal-Bench 4.0 & $0.87$ & $0.77$ & $0.82$ & $0.75$ \\
AutomationBench-AA & $0.81$ & $0.69$ & $0.79$ & $0.67$ \\
GDP.pdf & $0.69$ & $0.64$ & $0.74$ & $0.69$ \\
AA-Briefcase v1.1 & $0.58$ & $0.45$ & $0.55$ & $0.49$ \\
GDPval-AA v2.1 & $0.58$ & $0.52$ & $0.62$ & $0.50$ \\
CritPt & $0.56$ & $0.77$ & $0.64$ & $0.76$ \\
Humanity's Last Exam & $0.35$ & $0.48$ & $0.37$ & $0.52$ \\
AA-Omniscience & $0.32$ & $0.49$ & $0.32$ & $0.48$ \\
AA-LCR v1.1 & $0.26$ & $0.40$ & $0.26$ & $0.38$ \\
SciCode & $-0.03$ & $0.14$ & $-0.03$ & $0.13$ \\
\bottomrule
\end{tabular}
\end{table}

\section{Conclusion}

We introduce SepaRank, an evaluation protocol that measures models relative to each other. Each participant proposes open ended challenges answered by its peers, and is rewarded for separating the solvers' forecasts. The protocol does not require human-authored questions or a semantic judge, and is designed so that the difficulty of the challenges scales with the capabilities of the participating models.

We posit that the ability to find questions on which other capable systems disagree is indicative of intelligence, and remains measurable even when human examiners can no longer verify the posed challenges. The resulting rankings of SepaRank are consistent across code-constrained and open-ended games and correlate closely with established benchmarks of model ability.

Several directions remain open, such as a theory of the incentives that SepaRank creates, and in particular of the conditions under which reputation makes honest commitments optimal for proposers. Other separation objectives, such as entropy or divergence-based rewards, could replace the variance reward. Finally, the challenges and answers generated during play could serve as data for post-training and self-improvement.

\subsection*{AI use statement}

The LLMs evaluated in SepaRank generated challenges and answers as participants in the
evaluation protocol, so their outputs constitute the experimental
data. We used OpenAI Codex to assist with implementing and testing the experimental software, processing experimental records and external benchmark data, statistical analysis, preparing figures and tables, and formatting the paper in \LaTeX. We also used it to compose the technical low-level protocol/experiment description in the appendix. The authors take responsibility for
the final text, methods, results, and claims.

\subsection*{Ethics statement}

The challenges generated by the models under evaluation during the execution of SepaRank
may contain incorrect or biased content. Their usage as
experimental data does not endorse their claims.
The resulting rankings describe performance within the specified
protocol and are not presented as certification of a model's
safety, trustworthiness, or suitability for high-stakes deployment.

\subsection*{Reproducibility statement}

Section~\ref{sec:experiments} describes the experimental design and
reported measurements. Appendix~\ref{a-implementation} documents
the model identifiers, generation settings, random seeds, scoring
rules, information available to players, and handling of invalid
responses. Appendix~\ref{a-prompts} provides the prompts used in the
experiments.
Exact regeneration of model responses may vary with stochastic
sampling and changes to provider-hosted models.

\bibliographystyle{iclr2027_conference}
\bibliography{references}

\appendix
\newpage

\section{A look into authored challenges}
\label{a-qualitative}

The following examples illustrate the kinds of challenges produced in the
code-constrained and open-ended games. We retain the original question and program text, with
mathematical typography adjusted for readability. Throughout, $y$ denotes the proposer's committed bit $y^{(r)}_i$ and
$P$ the proposer reward $P^{(r)}_i$ of that match. 

\subsection{Open-ended questions}

\begin{table}[htbp]
\centering\small
\setlength{\tabcolsep}{4pt}
\caption{Selected open-ended questions. The column $p_j>0.5$ counts how many
of the 9 solvers assigned more than half their probability to
commitment 1.}
\label{tab:questions}
\begin{tabular}{@{}p{0.71\linewidth}crr@{}}
\toprule
Question & $y$ & $p_j>0.5$ & $P$ \\
\midrule
\textbf{Q1} As of December 31, 2024, was the Mersenne prime $2^{82{,}589{,}933}-1$ the largest prime number known to have been discovered? & 0 & 2/9 & 0.6379 \\
\addlinespace[5pt]
\textbf{Q2} Under the standard FIDE Laws of Chess, can a king legally capture an enemy piece if that piece is protected only by an enemy piece that is absolutely pinned to its own king? & 0 & 2/9 & 0.6206 \\
\addlinespace[5pt]
\textbf{Q3} Does the English word `antidisestablishmentarianism' (spelled in standard lowercase form) contain exactly four occurrences of the letter `t'? & 0 & 1/9 & 0.4104 \\
\addlinespace[5pt]
\textbf{Q4} Let $f:[0,1]\to\mathbb{R}$ be continuous on $[0,1]$ and differentiable with a finite derivative at every point of $(0,1)$. Suppose $f'(x)\ge0$ for Lebesgue-almost every $x\in(0,1)$, with no bound assumed on $f'$. Must $f$ be nondecreasing on $[0,1]$? & 1 & 5/9 & 0.8696 \\
\addlinespace[5pt]
\textbf{Q5} In the mathematical problem of tiling a square with finitely many smaller squares of pairwise distinct side lengths (a perfect squared square), is the smallest possible number of smaller squares exactly 21? & 0 & 9/9 & 0.0005 \\
\addlinespace[5pt]
\textbf{Q6} Is 0 a natural number? & 1 & 3/9 & 0.0195 \\
\bottomrule
\end{tabular}
\end{table}

\emph{Facts with a moving boundary.} Q1 asks for a familiar mathematical
record at a specific date. The old record had been surpassed by the
October 2024 discovery of $2^{136{,}279{,}841}-1$.\footnote{\href{https://www.mersenne.org/primes/?press=M136279841}{GIMPS discovery announcement}, 21 October 2024.}
Nevertheless, 2 solvers assigned more than half their probability to
commitment 1. The question was proposed by GPT-6 Astra in open-ended
game 1, round 1.

\emph{A familiar rule with an unfamiliar consequence.} Q2 turns on the
distinction between being able to move a piece and attacking a square.
Under the FIDE rules, a pinned piece still attacks the relevant square;
the king cannot capture into that attack.\footnote{\href{https://handbook.fide.com/chapter/e012023}{FIDE Laws of Chess}, Articles 3.1.3 and 3.9.1--3.9.2.}
GPT-5.6 Terra proposed this question in open-ended game 3, round 2.
The panel again split 2 versus 7 around $0.5$.

\emph{Character-level questions.} Q3 asks a small, deterministic question
about the spelling of a long word.
``Antidisestablishmentarianism'' contains 3 occurrences of
\texttt{t}, not 4. Claude Fable 5.1 proposed the question in open-ended
game 1, round 8. Of the 9 solvers, 8 reported probabilities below $0.5$ and the remaining solver
reported 1. The resulting reward, $P=0.4104$, illustrates why a headcount
alone does not determine separation, since confidence and solver weights also
matter.

\emph{Fine mathematical distinctions.} Q4 moves from checking a finite
object to distinguishing nearby theorem statements. It replaces an
everywhere condition on the derivative with an almost-everywhere
condition, while explicitly retaining differentiability at every
interior point. GPT-6 Astra proposed it in open-ended game 2, round 11.
Forecasts ranged from $0.005$ to $1$, with 5 above $0.5$ and 4 below,
giving $P=0.8696$.

\emph{Agreement despite a different commitment.} Q5 provides a useful
contrast to the high-separation examples. GPT-5.6 Terra committed 0 in
open-ended game 1, round 11, but every solver reported at least $0.97$.
The panel's near-agreement gave the proposer only $P=0.0005$, while its
mean Brier loss against the commitment was $0.9758$. Proposer reward and
solver loss therefore describe different aspects of the same exchange.
Agreement can produce little separation even when the forecasts are far
from the committed target.

\emph{Shared uncertainty.} Q6 illustrates a different route to a low
proposer reward. Muse Spark 1.2 asked ``Is 0 a natural number?'' in round 2
of the first anonymous author ablation game, committing 1. The 9 forecasts,
sorted for readability, were
\[
 0.35,\;0.50,\;0.50,\;0.50,\;0.50,\;0.50,\;0.55,\;0.60,\;0.60.
\]
Exactly 5 solvers reported $0.5$, and 8 were within $0.1$ of it.
With the uniform weights used during warmup, the proposer received only
$P=0.0195$. The reports clustered around shared uncertainty, producing
little separation. If all 9 solvers had reported exactly $0.5$, the
proposer reward would have been zero.

\subsection{Code-constrained questions}

The code-constrained games also produced short challenges whose difficulty lies in
language details rather than program length. The next 2 examples were
both recorded as returning 1 in the experiment's CPython 3.12.6 runtime,
and both proposers committed 1.

\emph{Floating-point summation.} Claude Fable 5.1 submitted the following
program in code-constrained game 5, round 8. It compares the built-in sum, repeated
addition, and \texttt{math.fsum} on the same 3 numbers.

\begin{promptbox}{P1: Three ways to add the same numbers}
import math

def main():
    values = [0.1, 0.2, 0.3]
    builtin_total = sum(values)
    loop_total = 0.0
    for v in values:
        loop_total += v
    exact = math.fsum(values)
    return int(builtin_total == 0.6 and loop_total != 0.6 and builtin_total == exact)
\end{promptbox}

Python 3.12 changed the floating-point summation algorithm used by
\texttt{sum}; it need not reproduce a loop of successive
additions.\footnote{Python 3.12 documentation for \href{https://docs.python.org/3.12/library/functions.html\#sum}{\texttt{sum}}.}
The panel split 4 versus 5 around $0.5$, with forecasts spanning
the full interval $[0,1]$, and the proposer received $P=0.9410$.

\emph{Equality and shared objects.} GPT-5.6 Terra submitted the next
program in code-constrained game 5, round 6.

\begin{promptbox}{P2: Unequal to itself, equal inside a dataclass}
import dataclasses

@dataclasses.dataclass
class Box:
    value: float

def main():
    x = float("nan")
    a = Box(x)
    b = Box(x)
    return int(x != x and a == b)
\end{promptbox}

The 2 boxes contain the same floating-point object. A NaN compares
unequal to itself, but the generated dataclass equality compares tuples
of fields, whose comparison treats identical contained objects as
equal.\footnote{Python 3.12 documentation for \href{https://docs.python.org/3.12/library/dataclasses.html}{dataclass equality} and
\href{https://docs.python.org/3.12/reference/expressions.html\#value-comparisons}{container comparisons}.}
Thus both tests in the return statement hold in the recorded runtime.
The panel again split 4 versus 5, with forecasts from 0 to 1,
giving $P=0.9657$.

\Needspace{19\baselineskip}
\emph{The same program with different commitments.} Claude Sonnet 5
submitted the following identical program in 2 games. Both executions
returned 0, but the recorded commitments differed.

\begin{promptbox}{P3: Parity of a sum of multiples}
def main():
    total = 0
    for i in range(1, 200000):
        if i % 3 == 0 or i % 5 == 0:
            total += i
    return total % 2
\end{promptbox}

\begin{table}[htbp]
\centering\small
\caption{The 2 submissions of P3. The last column is the unweighted mean
Brier loss of the 9 solver replies in that match.}
\label{tab:repeated-program}
\begin{tabular}{@{}lccrr@{}}
\toprule
Submission & $y$ & Executed bit & $P$ & Mean Brier \\
\midrule
Game 1, round 3 & 1 & 0 & 0.0110 & 0.9437 \\
Game 7, round 2 & 0 & 0 & 0.0010 & 0.0003 \\
\bottomrule
\end{tabular}
\end{table}

All 9 solvers assigned less than $0.5$ to commitment 1 in both
matches. The resulting separation was small in each case, but the
solver losses differed sharply. The commitment can therefore change
solver scores even when the program and its returned bit stay the same.

\newpage 

\section{Implementation parameters and information structure}
\label{a-implementation}

This appendix records the configuration used for the reported experiments.
The 2 main settings differ in the challenges that players can propose,
namely Python programs in the code-constrained games and yes/no questions in the open-ended
games. Both use the proposer's hidden commitment as the solver's target.
The complete prompts are reproduced in Appendix~\ref{a-prompts}.

\subsection{Games and model roster}

We ran 10 code-constrained games and 10 open-ended games, each with 10 players and
20 rounds. In every round, each player proposed once and the other 9
players were asked to solve each valid proposal. This gives 10 matches per
round. In total, the experiment contains 450
completed game-rounds and 4,500 matches.

Each game assigned the models pseudonyms that remained fixed within that
game. Model identities were not supplied to the players. The 5 anonymous author ablation games used the first
5 seeds and the same pseudonym assignments as their open-ended
counterparts. The anonymous author ablation uses only the first 10 rounds of
each paired game, so both sides have the same horizon.

\begin{table}[htbp]
\centering\small
\caption{Model roster and model identifiers recorded in the experiment.}
\label{tab:current-roster}
\begin{tabular}{@{}lll@{}}
\toprule
Provider & Model & API model identifier \\
\midrule
OpenAI & GPT-6 Astra & \texttt{gpt-6-astra} \\
& GPT-5.6 Terra & \texttt{gpt-5.6-terra} \\
Anthropic & Claude Fable 5.1 & \texttt{claude-fable-5-1} \\
& Claude Sonnet 5 & \texttt{claude-sonnet-5} \\
Google & Gemini 3.1 Pro Preview & \texttt{gemini-3.1-pro-preview} \\
& Gemini 3.8 Flash & \texttt{gemini-3.8-flash} \\
xAI & Grok 4.6 & \texttt{grok-4.6} \\
& Grok 4.5 & \texttt{grok-4.5} \\
Meta & Muse Spark 1.1 & \texttt{muse-spark-1.1} \\
& Muse Spark 1.2 & \texttt{muse-spark-1.2} \\
\bottomrule
\end{tabular}
\end{table}

All models used their provider's low reasoning-effort setting. Proposer requests allowed 12,288 output
tokens. 

\subsection{Scoring and solver weights}

Table~\ref{tab:params} gives the main parameters. Each model has 2 running
means, one over its proposer rewards and one over its solver Brier losses. Each scored
proposal contributes once to the first mean, and each scored solver reply
contributes once to the second. Both include the warmup rounds. A missing
observation is left unscored rather than entered as zero.

\begin{table}[htbp]
\centering\small
\caption{Shared protocol parameters.}
\label{tab:params}
\begin{tabular}{@{}p{0.36\linewidth}p{0.60\linewidth}@{}}
\toprule
Parameter & Setting \\
\midrule
Players per game & 10 \\
Matches per round & 10, with 1 proposer and 9 solvers each \\
Main-game horizon & 20 rounds \\
Anonymous author ablation horizon & 10 rounds; open-ended games only \\
Public history & Last 5 completed rounds, plus cumulative standings \\
Warmup & First 3 rounds; uniform solver weights \\
Weight update & After every completed round \\
Weight temperature & $0.25$ \\
Uniform weight mixture & $0.05$ \\
Minimum usable replies & $\max\{2,\lceil 0.5\times9\rceil\}=5$ \\
Proposer reward & 4 times the weighted population variance \\
Solver loss & Brier loss against the proposer's committed bit \\
Invalid-proposal reward & $-1$ \\
Malformed-reply loss & $1$, after one unsuccessful format retry \\
\bottomrule
\end{tabular}
\end{table}

During the warmup rounds $r\le 3$, $w^{(r)}_i=1/N$. After each completed round
$r\ge 3$, let $\bar L^{(r)}_i$ be player $i$'s mean Brier loss over all its scored
solver replies in rounds $1,\dots,r$. Round $r+1$ uses
\[
 w^{(r+1)}_i=0.95\,\frac{\exp(-\bar L^{(r)}_i/0.25)}{\sum_k\exp(-\bar L^{(r)}_k/0.25)}
       +\frac{0.05}{N},
\]
where the sum is over the $N$ active players. Uniform weights are also used
if any active player has no scored solver observations. Lower loss gives a solver more influence on the
proposer reward, while the uniform component keeps every active solver's
weight positive.

For proposal $q^{(r)}_i$, the proposer's own weight is excluded and the
remaining weights are renormalized over the set $U^{(r,i)}\subseteq[N]\setminus\{i\}$
of solvers with usable replies. Writing these
weights as $\tilde w^{(r,i)}_j=w^{(r)}_j/\sum_{k\in U^{(r,i)}}w^{(r)}_k$ and the corresponding forecasts as $p^{(r,i)}_j$, the reward is
\[
 \mu^{(r)}_i=\sum_{j\in U^{(r,i)}} \tilde w^{(r,i)}_j p^{(r,i)}_j, \qquad
 P^{(r)}_i=4\sum_{j\in U^{(r,i)}} \tilde w^{(r,i)}_j\bigl(p^{(r,i)}_j-\mu^{(r)}_i\bigr)^2.
\]
The joint score reported after play is
$J_i=(P_i+1-L_i)/2$, with $P_i=\bar P^{(R)}_i$ and $L_i=\bar L^{(R)}_i$. It is computed separately for code-constrained and
open-ended games and is not used to update weights or determine what any
player sees during play.

Reported scores average over games, weighting each game equally.
Score intervals are pointwise 95\% percentile intervals from 50,000
whole-game bootstrap resamples. Each resampled game retains all models
and both roles together. For the anonymous author ablation, resampling
retains the 5 matched game pairs. Its intervals enumerate all $5^5$
ordered bootstrap resamples of those pairs.

\subsection{What players see}

All matches in a round use the same round-start history, standings, and
weights. Results from that round are published together after every match
has finished. A turn therefore cannot see another player's current
commitment or report. Within a match, the commitment is fixed before the
solver replies are collected.

Every player sees the cumulative standings and the last 5 completed
rounds of full challenges, commitments, solver reports, and results.
Earlier rounds still contribute to the cumulative scores and weights.
Past proposers are identified by pseudonym. Each turn is otherwise a fresh
request, and no private conversation from earlier turns is supplied. A format
retry retains the preceding reply within that turn.

The anonymous author ablation removes
only the explicit attribution of the current proposer and informs proposers of
that change. It retains the attributed public history and cumulative
standings. Thus the ablation tests the value of the explicit proposer label,
not whether the proposer's identity can ever be inferred from content or history.

\subsection{Challenge validation and execution}

\emph{Code-constrained.} A submission contains a Python program and a separate final
\texttt{ANSWER: 0} or \texttt{ANSWER: 1} line. The parser selects the last
complete fenced block tagged \texttt{python}, \texttt{py}, or
\texttt{python3}. The program must be at most 8,000 characters and define
a top-level, synchronous \texttt{main()} with no arguments, returning an
actual integer 0 or 1. The program
must pass static checks and terminate in 2 independent executions with
the same returned bit.

The recorded runtime was CPython 3.12.6 on Darwin arm64, with
\texttt{PYTHONHASHSEED=0} and 64-bit \texttt{siphash13} hashing.
Each execution had a 30-second wall-clock limit and a 4-second total
process CPU limit, comprising 2 configured seconds and 2 seconds of
startup allowance. A 256 MiB address-space limit was configured. The import allowlist
and forbidden names and attributes are reproduced in
Appendix~\ref{a-prompts}. They exclude filesystem and network access,
dynamic execution, reflection, and private-state input. Class patterns
may use only permitted public keyword attributes, with no positional
class patterns. Execution establishes validity and records the returned
bit but does not replace the proposer's commitment as the scoring target.

\emph{Open-ended questions.} A submission is a JSON object containing a
question and a commitment in $\{0,1\}$. The question must be nonempty and
at most 2,000 characters after trimming outer whitespace. The prompt asks
for a yes/no question with one objectively resolvable answer shared by
all readers, independent of their identities, private states, and the
hidden commitment. The format, length, and binary
commitment are checked without a semantic judge.

\subsection{Parsing and failure rules}

A solver is asked for a JSON object with a finite numerical
\texttt{p\_one} field. The parser tolerates surrounding prose and JSON
fences, selects the last parsed object with that top-level field, and
gives JSON-fenced objects precedence. Finite values outside $[0,1]$ are
clipped to that interval. A parsing failure permits one format retry. If
a question commitment is expressed as an equivalent Boolean, binary
number, or recognized yes/no string, it is normalized to 0 or 1. If
a code-constrained submission contains a program but lacks a commitment, the retry
requests only the commitment and retains the program. A program that
fails static or execution checks receives no repair attempt.

An invalid proposal receives reward $-1$ and is included in the proposer's
mean. A solver reply that remains unparseable
after the retry receives Brier loss 1 when the target is available, but
does not enter the variance calculation. A valid proposal with fewer
than 5 usable replies is unscored for the proposer. Communication or
infrastructure failures are left unscored and never penalize a player.

At a round boundary, a player is held out if it has at least 6 cumulative
proposer or solver turns and at least half are failed turns. A holdout pauses the game before further rounds. No player met
this holdout rule in the reported experiment.

\newpage 
\section{Game prompts}
\label{a-prompts}

This appendix reproduces the prompts used in the reported experiment.
Each initial request is one user message that concatenates, in order, the
identity header, the fixed rules, the public history and standings, and
the applicable role block. Table~\ref{tab:prompt-blocks} identifies the blocks
used in each setting. Angle-bracket fields below stand for the actual
pseudonyms, history, standings, panel weights, or current challenge.
Fixed wording, including spacing, is retained as supplied to the models.

\begin{table}[htbp]
\centering\small
\caption{Prompt assembly. All settings use the shared identity header and
history scaffold.}
\label{tab:prompt-blocks}
\begin{tabular}{@{}p{0.28\linewidth}p{0.20\linewidth}p{0.44\linewidth}@{}}
\toprule
Setting & Fixed rules & Role block \\
\midrule
Code-constrained & Code & Code proposer or code solver \\
Open-ended & Open-ended & Open-ended proposer or solver \\
Anonymous author ablation & Open-ended & Corresponding author-withheld block \\
\bottomrule
\end{tabular}
\end{table}

\subsection{Shared header and fixed rules}

\begin{promptbox}{Shared identity header}
PLAYER: <PLAYER_PSEUDONYM>.
\end{promptbox}

\begin{promptbox}{Code: fixed rules shared by proposer and solver}
FULL-PANEL GAME: TWO CUMULATIVE SCORES
Active players N=10. Names remain fixed; use these assigned names.

ROLES AND SIMULTANEOUS ROUNDS
Every active player proposes one challenge per round. For proposer h,
S=active_players\{h}, |S|=9. Every member of S answers that challenge
separately, regardless of weight. You solve every other player's valid challenge.
There are N one-proposer matches per round, no pairs or byes. All N matches run
concurrently with the SAME round-start standings, weights and history. Publish scores and history only after ALL matches finish. Current
commitments and reports remain hidden until reports are locked.

TWO OBJECTIVES
Maximize your cumulative mean proposer reward and minimize your cumulative mean
solver Brier loss. The final evaluation will combine both components monotonically
and with equal importance: improving either component while holding the other
fixed cannot worsen your final score. The exact combination will be chosen in
post-processing. No combined score, z-score or overall ranking is used during play.
Each scored proposal contributes once to its arithmetic mean; each scored solver
response contributes once to its arithmetic mean. Unscored quantities do not
become fictitious zeros. No observations means an undefined mean.

PROPOSER REWARD
For a valid challenge, C is the subset of S with usable reports.
Required |C|>=q=max(2,ceil(0.5*9))=5.
Normalize frozen solver weights over C: w_j=W_j/sum_{k in C}W_k, uniform if zero mass.
  mu=sum_{j in C}w_j*p_j
  P=4*sum_{j in C}w_j*(p_j-mu)^2.
This is weighted population variance scaled to [0,1]. Repeated challenges are allowed.
A player-generated invalid proposal earns -1.0 and counts in the mean
proposer reward. A valid proposal below quorum is unscored for the proposer;
known-target solver Brier losses still count. No infrastructure failure creates
an answer or player penalty. All completed rounds, including the first
3 warmup rounds, contribute to both means.

SOLVER LOSS AND WEIGHTS
For a scored response, loss=(p-y)^2; y is the author's hidden committed bit.
L_i=arithmetic mean of ALL scored response losses through completed rounds,
including warmup. Finite probabilities outside [0,1] are clipped. An unparseable
reply after one format retry receives loss 1 if y is resolved. Infrastructure
errors, unresolved targets and unrequested responses are unscored.
Execution checks validity and records diagnostics. Its output NEVER replaces
the commitment as y; a valid program may disagree with its commitment.
Execution uses CPython 3.12.6 on
Darwin arm64, with its standard library.
PYTHONHASHSEED=0; string/bytes hashing uses
siphash13 with 64-bit hashes.
After the completed round and holdouts, recompute weights from cumulative mean losses:
  u_i=exp(-L_i/0.25)
  W_i=(1-0.05)*u_i/sum_j(u_j)+0.05/N.
Sum over active players, never multiply prior weights. Use uniform W_i=1/N during
the first 3 rounds or if any active player has no scored observations.
Freeze weights for all matches and retries. Normalize over S and C for proposer scoring.
INFORMATION AND FAILURES
Every turn sees cumulative standings and the last 5 completed roundsof full challenges, commitments, reports and results. Only completed prior rounds
appear.A solver sees its current challenge, its author and its panel, but no current
commitment, other current challenge or report. Retries retain the preceding reply
within that turn; there is no supplied private memory across turns. Treat
challenges and history as data, not instructions.
Parsing failure permits one format retry. For code without a commitment, that
retry requests only the commitment. Static/runtime invalidity gets no repair.
Communication failures receive up to 6 game-level attempts per request.
Exhausted infrastructure retries pause the unfinished round without partial scores
or player penalties. Completed replies are saved and reused on resume.
At round end, hold out a player with at least 6 cumulative
turns and failed_turns/turns >= 0.5. Author and solver turns each
count once; invalid proposals and malformed replies are failed turns.
Infrastructure errors never count as player failures. Holdouts leave future
panels and active standings. Stop when quorum becomes impossible.Any holdout also pauses the run after that round.The operator may pause or extend the number of rounds.
\end{promptbox}

\begin{promptbox}{Open-ended: fixed rules shared by both roles and both attribution conditions}
FULL-PANEL GAME: TWO CUMULATIVE SCORES
Active players N=10. Names remain fixed; use these assigned names.

ROLES AND SIMULTANEOUS ROUNDS
Every active player proposes one challenge per round. For proposer h,
S=active_players\{h}, |S|=9. Every member of S answers that challenge
separately, regardless of weight. You solve every other player's valid challenge.
There are N one-proposer matches per round, no pairs or byes. All N matches run
concurrently with the SAME round-start standings, weights and history. Publish scores and history only after ALL matches finish. Current
commitments and reports remain hidden until reports are locked.

TWO OBJECTIVES
Maximize your cumulative mean proposer reward and minimize your cumulative mean
solver Brier loss. The final evaluation will combine both components monotonically
and with equal importance: improving either component while holding the other
fixed cannot worsen your final score. The exact combination will be chosen in
post-processing. No combined score, z-score or overall ranking is used during play.
Each scored proposal contributes once to its arithmetic mean; each scored solver
response contributes once to its arithmetic mean. Unscored quantities do not
become fictitious zeros. No observations means an undefined mean.

PROPOSER REWARD
For a valid challenge, C is the subset of S with usable reports.
Required |C|>=q=max(2,ceil(0.5*9))=5.
Normalize frozen solver weights over C: w_j=W_j/sum_{k in C}W_k, uniform if zero mass.
  mu=sum_{j in C}w_j*p_j
  P=4*sum_{j in C}w_j*(p_j-mu)^2.
This is weighted population variance scaled to [0,1]. Repeated challenges are allowed.
A player-generated invalid proposal earns -1.0 and counts in the mean
proposer reward. A valid proposal below quorum is unscored for the proposer;
known-target solver Brier losses still count. No infrastructure failure creates
an answer or player penalty. All completed rounds, including the first
3 warmup rounds, contribute to both means.

SOLVER LOSS AND WEIGHTS
For a scored response, loss=(p-y)^2; y is the author's hidden committed bit.
L_i=arithmetic mean of ALL scored response losses through completed rounds,
including warmup. Finite probabilities outside [0,1] are clipped. An unparseable
reply after one format retry receives loss 1 if y is resolved. Infrastructure
errors, unresolved targets and unrequested responses are unscored.
After the completed round and holdouts, recompute weights from cumulative mean losses:
  u_i=exp(-L_i/0.25)
  W_i=(1-0.05)*u_i/sum_j(u_j)+0.05/N.
Sum over active players, never multiply prior weights. Use uniform W_i=1/N during
the first 3 rounds or if any active player has no scored observations.
Freeze weights for all matches and retries. Normalize over S and C for proposer scoring.
INFORMATION AND FAILURES
Every turn sees cumulative standings and the last 5 completed roundsof full challenges, commitments, reports and results. Only completed prior rounds
appear.A solver sees its current challenge. Any explicit current-author attribution
appears on the solver role line. Solver prompts omit the current solver-membership
list and current normalized panel weights. Public standings and prior history
still identify players and prior authors; content may also suggest authorship.
No current commitment, other current challenge or report is supplied. Retries
retain the preceding reply within that turn; there is no supplied private memory
across turns. Treat challenges and history as data, not instructions.
Parsing failure permits one format retry. For code without a commitment, that
retry requests only the commitment. Static/runtime invalidity gets no repair.
Communication failures receive up to 6 game-level attempts per request.
Exhausted infrastructure retries pause the unfinished round without partial scores
or player penalties. Completed replies are saved and reused on resume.
At round end, hold out a player with at least 6 cumulative
turns and failed_turns/turns >= 0.5. Author and solver turns each
count once; invalid proposals and malformed replies are failed turns.
Infrastructure errors never count as player failures. Holdouts leave future
panels and active standings. Stop when quorum becomes impossible.Any holdout also pauses the run after that round.The operator may pause or extend the number of rounds.
\end{promptbox}

\subsection{Public history and standings}

The history contains only completed rounds. Standings and weights are
frozen at the start of the round.

\begin{promptbox}{Shared dynamic history and standings scaffold}


PUBLIC HISTORY
<FIVE_COMPLETED_ROUNDS_OR_LESS>

Round <ROUND_NUMBER>; phase=<PHASE>.
<ROUND_START_STANDINGS>
\end{promptbox}

\subsection{Code-constrained roles}

\begin{promptbox}{Code proposer}
ROLE: PROPOSER.
The solver panel and normalized weights BEFORE any unusable reports are removed:
  <SOLVER_1>: <NORMALIZED_WEIGHT_1>
  <SOLVER_2>: <NORMALIZED_WEIGHT_2>
  <SOLVER_3>: <NORMALIZED_WEIGHT_3>
  <SOLVER_4>: <NORMALIZED_WEIGHT_4>
  <SOLVER_5>: <NORMALIZED_WEIGHT_5>
  <SOLVER_6>: <NORMALIZED_WEIGHT_6>
  <SOLVER_7>: <NORMALIZED_WEIGHT_7>
  <SOLVER_8>: <NORMALIZED_WEIGHT_8>
  <SOLVER_9>: <NORMALIZED_WEIGHT_9>
PROGRAM SCOPE: You may choose any computational topic, subject to the Python
and validity requirements below.
HISTORY REMINDER: PUBLIC HISTORY shows the full text and recorded committed answers
of all programs posed by any proposer in the retained prior sweeps.
Every solver in this match sees this same history, including your own past proposals.
Submit one self-contained Python program and commit its output bit. Define a
top-level synchronous main() with no arguments, returning an actual int 0 or 1.
Solvers receive your program and identity; the commitment remains hidden.
Validity requirements:
- At most 8000 characters.
- Per execution: 4 total process CPU seconds
  (2 configured + 2 startup allowance), 30 wall seconds,
  256 MiB address-space limit where supported by the host.
- Allowed imports: abc, array, bisect, cmath, collections, collections.abc, copy, dataclasses, decimal, fractions, functools, hashlib, heapq, itertools, math, operator, re, statistics, string, struct, unicodedata.
- No relative/wildcard imports or private/forbidden imported symbols.
- No private/dunder attribute access. Forbidden names: __builtins__, __import__, __loader__, __spec__, breakpoint, compile, delattr, dir, eval, exec, getattr, globals, hasattr, help, input, locals, open, setattr, vars.
- Forbidden attributes: Formatter, __builtins__, __import__, __loader__, __spec__, atexit, attrgetter, breakpoint, builtins, concurrent, ctypes, datetime, delattr, dir, eval, exec, faulthandler, format, format_map, ftplib, gc, get_field, getattr, globals, hasattr, help, http, importlib, input, inspect, io, locals, marshal, methodcaller, mro, multiprocessing, open, os, pathlib, pickle, random, requests, resource, secrets, setattr, shutil, signal, singledispatch, singledispatchmethod, socket, subprocess, sys, tempfile, threading, time, urllib, vars.
- Class patterns may use public, permitted keyword attributes only; no positional class patterns.
- No dynamic execution, reflection, filesystem/network I/O or private-state input.
- Two independent executions must terminate successfully with the same integer bit.
The last complete fenced block tagged python, py, or python3 is the submitted
program. Earlier blocks and surrounding prose are ignored. The standalone final
ANSWER line commits to that submitted program.
Use this format:
```python
<program>
```
ANSWER: <0 or 1>
\end{promptbox}

\begin{promptbox}{Code solver}
ROLE: SOLVER. Current proposer: <AUTHOR_PSEUDONYM>.
The solver panel and normalized weights BEFORE any unusable reports are removed:
  <SOLVER_1>: <NORMALIZED_WEIGHT_1>
  <SOLVER_2>: <NORMALIZED_WEIGHT_2>
  <SOLVER_3>: <NORMALIZED_WEIGHT_3>
  <SOLVER_4>: <NORMALIZED_WEIGHT_4>
  <SOLVER_5>: <NORMALIZED_WEIGHT_5>
  <SOLVER_6>: <NORMALIZED_WEIGHT_6>
  <SOLVER_7>: <NORMALIZED_WEIGHT_7>
  <SOLVER_8>: <NORMALIZED_WEIGHT_8>
  <SOLVER_9>: <NORMALIZED_WEIGHT_9>
HISTORY REMINDER: PUBLIC HISTORY shows the full text and recorded committed answers
of all programs posed by any proposer in the retained prior sweeps.
Every solver in this match sees this same history, including your own past proposals.
CHALLENGE
```python
<CURRENT_PROGRAM>
```

END CHALLENGE
Report p=P(y=1) under the resolution rule above. Return exactly one JSON object,
with a finite number in [0,1], without surrounding text or fences:
{"p_one": <probability>}
Surrounding prose or JSON fences are tolerated; only the extracted p_one is scored.
Use the last parsed object with a top-level p_one field; JSON-fenced objects take precedence.
\end{promptbox}

\subsection{Open-ended roles}

\begin{promptbox}{Open-ended proposer}
ROLE: PROPOSER.
The solver panel and normalized weights BEFORE any unusable reports are removed:
  <SOLVER_1>: <NORMALIZED_WEIGHT_1>
  <SOLVER_2>: <NORMALIZED_WEIGHT_2>
  <SOLVER_3>: <NORMALIZED_WEIGHT_3>
  <SOLVER_4>: <NORMALIZED_WEIGHT_4>
  <SOLVER_5>: <NORMALIZED_WEIGHT_5>
  <SOLVER_6>: <NORMALIZED_WEIGHT_6>
  <SOLVER_7>: <NORMALIZED_WEIGHT_7>
  <SOLVER_8>: <NORMALIZED_WEIGHT_8>
  <SOLVER_9>: <NORMALIZED_WEIGHT_9>
QUESTION SCOPE: You may ask any yes/no question in any knowledge domain,
subject to the validity requirements below.
HISTORY REMINDER: PUBLIC HISTORY shows the full text and recorded committed answers
of all questions posed by any proposer in the retained prior sweeps.
Every solver in this match sees this same history, including your own past proposals.
Submit one yes/no question and commit its answer bit (0=no, 1=yes). The question must have a
single objectively resolvable answer shared by all readers; it must not depend
on a reader's identity, private state, or this hidden commitment. Solvers receive
your question and your assigned name; the commitment remains hidden until reports lock.
Prior public history retains the recorded author names in both conditions.
Validity requires a nonempty question of at most 2000 characters
(after trimming outer whitespace) and a commitment in {0,1}. Semantic correctness
is not machine-verified.
Return exactly one JSON object, without surrounding text or fences:
{"question": "<question>", "answer": <0 or 1>}
\end{promptbox}

\begin{promptbox}{Open-ended solver}
ROLE: SOLVER. Current proposer: <AUTHOR_PSEUDONYM>.
HISTORY REMINDER: PUBLIC HISTORY shows the full text and recorded committed answers
of all questions posed by any proposer in the retained prior sweeps.
Every solver in this match sees this same history, including your own past proposals.
CHALLENGE
<CURRENT_QUESTION>

END CHALLENGE
Report p=P(y=1) under the resolution rule above. Return exactly one JSON object,
with a finite number in [0,1], without surrounding text or fences:
{"p_one": <probability>}
Surrounding prose or JSON fences are tolerated; only the extracted p_one is scored.
Use the last parsed object with a top-level p_one field; JSON-fenced objects take precedence.
\end{promptbox}

\subsection{Anonymous author ablation}

\begin{promptbox}{Open-ended proposer: author withheld (ablation only)}
ROLE: PROPOSER.
The solver panel and normalized weights BEFORE any unusable reports are removed:
  <SOLVER_1>: <NORMALIZED_WEIGHT_1>
  <SOLVER_2>: <NORMALIZED_WEIGHT_2>
  <SOLVER_3>: <NORMALIZED_WEIGHT_3>
  <SOLVER_4>: <NORMALIZED_WEIGHT_4>
  <SOLVER_5>: <NORMALIZED_WEIGHT_5>
  <SOLVER_6>: <NORMALIZED_WEIGHT_6>
  <SOLVER_7>: <NORMALIZED_WEIGHT_7>
  <SOLVER_8>: <NORMALIZED_WEIGHT_8>
  <SOLVER_9>: <NORMALIZED_WEIGHT_9>
QUESTION SCOPE: You may ask any yes/no question in any knowledge domain,
subject to the validity requirements below.
HISTORY REMINDER: PUBLIC HISTORY shows the full text and recorded committed answers
of all questions posed by any proposer in the retained prior sweeps.
Every solver in this match sees this same history, including your own past proposals.
Submit one yes/no question and commit its answer bit (0=no, 1=yes). The question must have a
single objectively resolvable answer shared by all readers; it must not depend
on a reader's identity, private state, or this hidden commitment. Solvers receive
your question with the explicit current-author label withheld; the commitment remains hidden until reports lock.
Prior public history retains the recorded author names in both conditions.
Validity requires a nonempty question of at most 2000 characters
(after trimming outer whitespace) and a commitment in {0,1}. Semantic correctness
is not machine-verified.
Return exactly one JSON object, without surrounding text or fences:
{"question": "<question>", "answer": <0 or 1>}
\end{promptbox}

\begin{promptbox}{Open-ended solver: author withheld (ablation only)}
ROLE: SOLVER. Current proposer: withheld.
HISTORY REMINDER: PUBLIC HISTORY shows the full text and recorded committed answers
of all questions posed by any proposer in the retained prior sweeps.
Every solver in this match sees this same history, including your own past proposals.
CHALLENGE
<CURRENT_QUESTION>

END CHALLENGE
Report p=P(y=1) under the resolution rule above. Return exactly one JSON object,
with a finite number in [0,1], without surrounding text or fences:
{"p_one": <probability>}
Surrounding prose or JSON fences are tolerated; only the extracted p_one is scored.
Use the last parsed object with a top-level p_one field; JSON-fenced objects take precedence.
\end{promptbox}

\subsection{Format-retry instructions}

A format retry appends the preceding assistant output (or the literal
\texttt{(no output)}) and the applicable instruction below. It retains
the same history and round-start state. No separate reasoning field is
requested in the response JSON.

\begin{promptbox}{Malformed code author response}
Your previous reply did not match the required format. Respond now with exactly one fenced ```python code block defining main(), then a final line that is EXACTLY `ANSWER: 0` or `ANSWER: 1`. Nothing else.
\end{promptbox}

\begin{promptbox}{Code author: missing commitment only}
Your reply was missing the final commitment line. Reply now with ONLY one line, EXACTLY `ANSWER: 0` or `ANSWER: 1` — your claim of what main() returns.
\end{promptbox}

\begin{promptbox}{Malformed open-ended author response}
Your previous reply did not parse. Respond now with ONLY a JSON object on one line and nothing else: {"question": "<your yes/no question>", "answer": 0 or 1}
\end{promptbox}

\begin{promptbox}{Malformed solver response}
Reply with ONLY a JSON object on one line and nothing else: {"p_one": <number between 0 and 1>}
\end{promptbox}

\newpage
\section{Additional Experiments}
\label{a-additional_experiments}

\subsection{Dynamics}
\label{subsec:dynamics}
Figure~\ref{fig:trajectories} shows each model's cumulative proposer and solver scores through successive rounds, averaged over all 10 games. GPT-6 Astra's proposer advantage is visible from the start, while several competitors gain ground as play continues. Claude Fable 5.1's mean proposer reward rises from $0.147$ at round 5 to $0.401$ at round 20 in code-constrained games, and from $0.172$ to $0.424$ in open-ended games. The open-ended solver skill declines over the same period for every model, alongside increasing proposer rewards.

\begin{figure}[!htbp]
\centering
\begin{tikzpicture}
\begin{axis}[sr/base,width=150pt,height=93pt,at={(0pt,0pt)},anchor=north west,
xmin=.5,xmax=20.5,ymin=-0.13,ymax=0.73,xtick={1,5,10,15,20},ytick={0,.2,.4,.6},xmajorgrids=false,
ylabel={Proposer reward $\bar P^{(r)}$},
title={Code-constrained},
]
\path[fill=black!9] (axis cs:.5,-0.13) rectangle (axis cs:3.5,0.73);
\addplot[color=srModel0,solid,line width=.85pt,mark=*,mark size=1.55pt,mark repeat=4] coordinates {
(1,0.5945916049) (2,0.5577711111) (3,0.5302324082) (4,0.5430271508) (5,0.5693646966) (6,0.5906714874) (7,0.5699334016) (8,0.5702311752) (9,0.5686804732) (10,0.5899126227) (11,0.5906589222) (12,0.6009016515) (13,0.5961990127) (14,0.5968616405) (15,0.6093008556) (16,0.6136307034) (17,0.6267131384) (18,0.6260684445) (19,0.6238122344) (20,0.6285700459)
};
\addplot[color=srModel1,solid,line width=.85pt,mark=square*,mark size=1.55pt,mark repeat=4] coordinates {
(1,0.0729228318) (2,0.1444487577) (3,0.1604397479) (4,0.2270084170) (5,0.2822690936) (6,0.3244088937) (7,0.3138110772) (8,0.2925409259) (9,0.3200710539) (10,0.3195089389) (11,0.3453362330) (12,0.3498172291) (13,0.3580483803) (14,0.3576176493) (15,0.3596338280) (16,0.3653914806) (17,0.3702287995) (18,0.3648871270) (19,0.3598210692) (20,0.3637415741)
};
\addplot[color=srModel2,solid,line width=.85pt,mark=triangle*,mark size=1.55pt,mark repeat=4] coordinates {
(1,-0.0217629630) (2,-0.1013150617) (3,0.0060296296) (4,0.1001118949) (5,0.1472742741) (6,0.2269148576) (7,0.2862244134) (8,0.3162432640) (9,0.3226224701) (10,0.3449904964) (11,0.3763100734) (12,0.3981978332) (13,0.3980970076) (14,0.3947026030) (15,0.4066642794) (16,0.4075678032) (17,0.3982006447) (18,0.4028417015) (19,0.4067901498) (20,0.4010737793)
};
\addplot[color=srModel3,solid,line width=.85pt,mark=diamond*,mark size=1.55pt,mark repeat=4] coordinates {
(1,0.0603377778) (2,0.0527960494) (3,0.0663776132) (4,0.0777772942) (5,0.0874479178) (6,0.1462518954) (7,0.1460416052) (8,0.1416417977) (9,0.1472218032) (10,0.1488729276) (11,0.1481135324) (12,0.1460535005) (13,0.1262103665) (14,0.1241514166) (15,0.1253068695) (16,0.1290318063) (17,0.1254155095) (18,0.1295943909) (19,0.1259316552) (20,0.1415014331)
};
\addplot[color=srModel4,solid,line width=.85pt,mark=triangle,mark size=1.55pt,mark repeat=4] coordinates {
(1,0.0124345679) (2,-0.0193688889) (3,0.0112253498) (4,0.0267005609) (5,0.0549747400) (6,0.0671669462) (7,0.0771722377) (8,0.0721244170) (9,0.0759790487) (10,0.0745212345) (11,0.0731868491) (12,0.0671462725) (13,0.0692127524) (14,0.0693568835) (15,0.0682210130) (16,0.0577402021) (17,0.0600786789) (18,0.0567733092) (19,0.0618900367) (20,0.0566507978)
};
\addplot[color=srModel5,dashed,line width=.85pt,mark=+,mark size=1.55pt,mark repeat=4] coordinates {
(1,0.0941964082) (2,0.1231238831) (3,0.1408335352) (4,0.1239282360) (5,0.0979083583) (6,0.1753507773) (7,0.1719388751) (8,0.1876051345) (9,0.2132420743) (10,0.2064542894) (11,0.2236357734) (12,0.2270639859) (13,0.2411047916) (14,0.2614001239) (15,0.2700503493) (16,0.2751236614) (17,0.2768838307) (18,0.2752317907) (19,0.2776099810) (20,0.2739101481)
};
\addplot[color=srModel6,dashed,line width=.85pt,mark=x,mark size=1.55pt,mark repeat=4] coordinates {
(1,0.0246518519) (2,0.0782668364) (3,0.1066997807) (4,0.1287227402) (5,0.1115287611) (6,0.1278181732) (7,0.1587510995) (8,0.1672224900) (9,0.1800015437) (10,0.1956416053) (11,0.2065078149) (12,0.2181373137) (13,0.2343623358) (14,0.2480692386) (15,0.2575294454) (16,0.2572781532) (17,0.2682094650) (18,0.2712433559) (19,0.2652549385) (20,0.2668482558)
};
\addplot[color=srModel7,dashed,line width=.85pt,mark=square,mark size=1.55pt,mark repeat=4] coordinates {
(1,0.0460387018) (2,0.0867126003) (3,0.1019901286) (4,0.1066518099) (5,0.1211889057) (6,0.1580906092) (7,0.1819264755) (8,0.2037127695) (9,0.2205434786) (10,0.2318556163) (11,0.2312958084) (12,0.2409800491) (13,0.2535850530) (14,0.2484715824) (15,0.2500669153) (16,0.2514237126) (17,0.2503991419) (18,0.2504086426) (19,0.2537751086) (20,0.2544931133)
};
\addplot[color=srModel8,dashed,line width=.85pt,mark=diamond,mark size=1.55pt,mark repeat=4] coordinates {
(1,0.0352898765) (2,0.0708059259) (3,0.1168827588) (4,0.1509146991) (5,0.1894425170) (6,0.1829627331) (7,0.2036025056) (8,0.2010857064) (9,0.1774804017) (10,0.1881936607) (11,0.2010395164) (12,0.2010848534) (13,0.2253089355) (14,0.2235058143) (15,0.2359289675) (16,0.2368922504) (17,0.2372306186) (18,0.2369197836) (19,0.2371609390) (20,0.2375652576)
};
\addplot[color=srModel9,dashed,line width=.85pt,mark=star,mark size=1.55pt,mark repeat=4] coordinates {
(1,0.0503170370) (2,0.0999647111) (3,0.1668301366) (4,0.1428188453) (5,0.1583763434) (6,0.1576504783) (7,0.1913843564) (8,0.2054698321) (9,0.2129289500) (10,0.2206310449) (11,0.2371097116) (12,0.2250489046) (13,0.2169856843) (14,0.2283295874) (15,0.2375881100) (16,0.2507421032) (17,0.2470181105) (18,0.2414104576) (19,0.2426928702) (20,0.2362253706)
};
\end{axis}
\begin{axis}[sr/base,width=150pt,height=93pt,at={(184pt,0pt)},anchor=north west,
xmin=.5,xmax=20.5,ymin=-0.13,ymax=0.73,xtick={1,5,10,15,20},ytick={0,.2,.4,.6},xmajorgrids=false,
title={Open-ended},
]
\path[fill=black!9] (axis cs:.5,-0.13) rectangle (axis cs:3.5,0.73);
\addplot[color=srModel0,solid,line width=.85pt,mark=*,mark size=1.55pt,mark repeat=4] coordinates {
(1,0.4737690568) (2,0.4109708988) (3,0.4685916280) (4,0.4806163720) (5,0.4766464270) (6,0.4987039737) (7,0.5057090359) (8,0.5063181409) (9,0.4996881373) (10,0.5157825854) (11,0.5333085514) (12,0.5519132606) (13,0.5555393501) (14,0.5554190998) (15,0.5602012315) (16,0.5597699282) (17,0.5594287214) (18,0.5663092085) (19,0.5649254281) (20,0.5675504649)
};
\addplot[color=srModel1,solid,line width=.85pt,mark=square*,mark size=1.55pt,mark repeat=4] coordinates {
(1,0.0021111590) (2,0.1120267252) (3,0.1293630612) (4,0.1345100183) (5,0.1700817956) (6,0.2028296642) (7,0.2152569093) (8,0.2208778128) (9,0.2281836509) (10,0.2360494772) (11,0.2365021829) (12,0.2507378399) (13,0.2577345752) (14,0.2638076326) (15,0.2829141913) (16,0.2970237897) (17,0.2988526203) (18,0.3097800061) (19,0.3144067349) (20,0.3160527447)
};
\addplot[color=srModel2,solid,line width=.85pt,mark=triangle*,mark size=1.55pt,mark repeat=4] coordinates {
(1,0.0702205728) (2,0.0982722469) (3,0.1221821399) (4,0.1218021099) (5,0.1720301122) (6,0.1910313125) (7,0.2425865167) (8,0.2695580211) (9,0.2733604451) (10,0.2898068612) (11,0.3086464105) (12,0.3292337746) (13,0.3192927047) (14,0.3266379090) (15,0.3580023661) (16,0.3766037156) (17,0.3943828307) (18,0.4111262098) (19,0.4204842713) (20,0.4239321425)
};
\addplot[color=srModel3,solid,line width=.85pt,mark=diamond*,mark size=1.55pt,mark repeat=4] coordinates {
(1,0.0006159210) (2,0.0004878864) (3,0.0003603619) (4,0.0003437436) (5,0.0003196244) (6,0.0170907228) (7,0.0146603325) (8,0.0201539547) (9,0.0223976069) (10,0.0311001227) (11,0.0323428974) (12,0.0396192592) (13,0.0431696617) (14,0.0402651084) (15,0.0378302053) (16,0.0358654055) (17,0.0429338363) (18,0.0442623317) (19,0.0475652942) (20,0.0541030115)
};
\addplot[color=srModel4,solid,line width=.85pt,mark=triangle,mark size=1.55pt,mark repeat=4] coordinates {
(1,0.0768205827) (2,0.0877837333) (3,0.0843358881) (4,0.0775967318) (5,0.0621491374) (6,0.0675118896) (7,0.0667051760) (8,0.0691110576) (9,0.0731638392) (10,0.0697372036) (11,0.0711554028) (12,0.0756252993) (13,0.0797585982) (14,0.0850087215) (15,0.0860903850) (16,0.0908357351) (17,0.0958168895) (18,0.0942345946) (19,0.0973906320) (20,0.0962268773)
};
\addplot[color=srModel5,dashed,line width=.85pt,mark=+,mark size=1.55pt,mark repeat=4] coordinates {
(1,0.0034159506) (2,0.0725986914) (3,0.0940784329) (4,0.0928142802) (5,0.1238710719) (6,0.1418914497) (7,0.1446786977) (8,0.1532721115) (9,0.1473779234) (10,0.1578658901) (11,0.1741641125) (12,0.1856876152) (13,0.1853561192) (14,0.2039697075) (15,0.2175832442) (16,0.2335652854) (17,0.2456482052) (18,0.2465771529) (19,0.2512602555) (20,0.2615579693)
};
\addplot[color=srModel6,dashed,line width=.85pt,mark=x,mark size=1.55pt,mark repeat=4] coordinates {
(1,0.0025398025) (2,0.0397643210) (3,0.0267198025) (4,0.0201275781) (5,0.0515836817) (6,0.0708145203) (7,0.0646068804) (8,0.1001901238) (9,0.1033964785) (10,0.1208441497) (11,0.1316409014) (12,0.1399125171) (13,0.1522327383) (14,0.1581444668) (15,0.1619226230) (16,0.1649080398) (17,0.1684007558) (18,0.1795489251) (19,0.1942542883) (20,0.2103519379)
};
\addplot[color=srModel7,dashed,line width=.85pt,mark=square,mark size=1.55pt,mark repeat=4] coordinates {
(1,0.0066270222) (2,0.0394355014) (3,0.0518608954) (4,0.0451378773) (5,0.0809775814) (6,0.1235870502) (7,0.1113000491) (8,0.1062647212) (9,0.1276880135) (10,0.1351875262) (11,0.1289711240) (12,0.1337002888) (13,0.1321914860) (14,0.1401796893) (15,0.1485306970) (16,0.1552212245) (17,0.1628044729) (18,0.1625402968) (19,0.1689060871) (20,0.1833963756)
};
\addplot[color=srModel8,dashed,line width=.85pt,mark=diamond,mark size=1.55pt,mark repeat=4] coordinates {
(1,0.0969253630) (2,0.1342960889) (3,0.1025949432) (4,0.0914192061) (5,0.1043916982) (6,0.1491954887) (7,0.1668813403) (8,0.1711083554) (9,0.1778292009) (10,0.1846435302) (11,0.1934208865) (12,0.1966148617) (13,0.2033424751) (14,0.2086487481) (15,0.2097190549) (16,0.2139192753) (17,0.2189771622) (18,0.2282633689) (19,0.2358249354) (20,0.2413426598)
};
\addplot[color=srModel9,dashed,line width=.85pt,mark=star,mark size=1.55pt,mark repeat=4] coordinates {
(1,0.0398618074) (2,0.0826228540) (3,0.0833673052) (4,0.0891364773) (5,0.1076057314) (6,0.1247948099) (7,0.1232049958) (8,0.1505376020) (9,0.1747017668) (10,0.1839324470) (11,0.1751112474) (12,0.1866084003) (13,0.1943541445) (14,0.1939027802) (15,0.2027796194) (16,0.2013743657) (17,0.2057903668) (18,0.2195970126) (19,0.2379253859) (20,0.2437459283)
};
\end{axis}
\begin{axis}[sr/base,width=150pt,height=93pt,at={(0pt,-124pt)},anchor=north west,
xmin=.5,xmax=20.5,ymin=0.6,ymax=1.01,xtick={1,5,10,15,20},ytick={.6,.7,.8,.9,1},xmajorgrids=false,
ylabel={Solver skill $1-\bar L^{(r)}$},
xlabel={Completed round},
]
\path[fill=black!9] (axis cs:.5,0.6) rectangle (axis cs:3.5,1.01);
\addplot[color=srModel0,solid,line width=.85pt,mark=*,mark size=1.55pt,mark repeat=4] coordinates {
(1,0.8865263888) (2,0.8916806245) (3,0.8661881728) (4,0.8720940778) (5,0.8782381279) (6,0.8813548834) (7,0.8785496912) (8,0.8839926919) (9,0.8830174284) (10,0.8858016657) (11,0.8868219908) (12,0.8890556323) (13,0.8893951209) (14,0.8906813129) (15,0.8903019168) (16,0.8944589041) (17,0.8942404044) (18,0.8960770124) (19,0.8959895230) (20,0.8950125329)
};
\addplot[color=srModel1,solid,line width=.85pt,mark=square*,mark size=1.55pt,mark repeat=4] coordinates {
(1,0.8565496923) (2,0.8529221796) (3,0.8245067858) (4,0.8137979381) (5,0.8183804245) (6,0.8015108233) (7,0.8003285821) (8,0.7927636031) (9,0.7866421390) (10,0.7749152807) (11,0.7728892943) (12,0.7707817658) (13,0.7641262557) (14,0.7647509523) (15,0.7624171968) (16,0.7600527893) (17,0.7589018189) (18,0.7601886788) (19,0.7582963437) (20,0.7532043777)
};
\addplot[color=srModel2,solid,line width=.85pt,mark=triangle*,mark size=1.55pt,mark repeat=4] coordinates {
(1,0.8811288889) (2,0.8704995752) (3,0.8596965670) (4,0.8677643413) (5,0.8718568182) (6,0.8713859050) (7,0.8666956650) (8,0.8698750221) (9,0.8688905692) (10,0.8679019895) (11,0.8706355757) (12,0.8724720976) (13,0.8728638844) (14,0.8748482059) (15,0.8753905497) (16,0.8771502154) (17,0.8781772680) (18,0.8788162297) (19,0.8796338930) (20,0.8787268437)
};
\addplot[color=srModel3,solid,line width=.85pt,mark=diamond*,mark size=1.55pt,mark repeat=4] coordinates {
(1,0.8211065722) (2,0.7923966085) (3,0.7667860864) (4,0.7624798077) (5,0.7473784560) (6,0.7368041470) (7,0.7233415463) (8,0.7137640490) (9,0.7027142366) (10,0.7013339762) (11,0.6867305097) (12,0.6839589381) (13,0.6801135123) (14,0.6768703638) (15,0.6715582965) (16,0.6696049222) (17,0.6660988578) (18,0.6647855672) (19,0.6611707266) (20,0.6610228178)
};
\addplot[color=srModel4,solid,line width=.85pt,mark=triangle,mark size=1.55pt,mark repeat=4] coordinates {
(1,0.8687840706) (2,0.8693157534) (3,0.8405312530) (4,0.8300537065) (5,0.8229237895) (6,0.7990722969) (7,0.7800502363) (8,0.7704693746) (9,0.7553568080) (10,0.7432285197) (11,0.7329120229) (12,0.7300264274) (13,0.7220016740) (14,0.7160624658) (15,0.7139119927) (16,0.7091336652) (17,0.7094042724) (18,0.7063229188) (19,0.7053012251) (20,0.7010400060)
};
\addplot[color=srModel5,dashed,line width=.85pt,mark=+,mark size=1.55pt,mark repeat=4] coordinates {
(1,0.8364101389) (2,0.8444582843) (3,0.8324143483) (4,0.8280969951) (5,0.8209906224) (6,0.8240033022) (7,0.8143535275) (8,0.8157192364) (9,0.8121275852) (10,0.8123930901) (11,0.8161290430) (12,0.8135347515) (13,0.8124094434) (14,0.8124059771) (15,0.8121107645) (16,0.8111658542) (17,0.8126484293) (18,0.8134414067) (19,0.8175333341) (20,0.8194958064)
};
\addplot[color=srModel6,dashed,line width=.85pt,mark=x,mark size=1.55pt,mark repeat=4] coordinates {
(1,0.7706997667) (2,0.7809106935) (3,0.7665035137) (4,0.7660666452) (5,0.7676704901) (6,0.7484592188) (7,0.7417710028) (8,0.7444020026) (9,0.7411622124) (10,0.7436477337) (11,0.7417671476) (12,0.7458656111) (13,0.7502119923) (14,0.7535872572) (15,0.7507949390) (16,0.7502234949) (17,0.7512905554) (18,0.7528368434) (19,0.7575063215) (20,0.7589588572)
};
\addplot[color=srModel7,dashed,line width=.85pt,mark=square,mark size=1.55pt,mark repeat=4] coordinates {
(1,0.7767570833) (2,0.7870319935) (3,0.7793467749) (4,0.7563758316) (5,0.7705157008) (6,0.7517812080) (7,0.7528460102) (8,0.7488074218) (9,0.7488691655) (10,0.7490247561) (11,0.7496447628) (12,0.7505159269) (13,0.7549304778) (14,0.7526130222) (15,0.7496188707) (16,0.7499493841) (17,0.7497478429) (18,0.7476201264) (19,0.7496444568) (20,0.7507337843)
};
\addplot[color=srModel8,dashed,line width=.85pt,mark=diamond,mark size=1.55pt,mark repeat=4] coordinates {
(1,0.8025352639) (2,0.8196687734) (3,0.8105755179) (4,0.7900805807) (5,0.7950921767) (6,0.7705476009) (7,0.7673705036) (8,0.7609804187) (9,0.7547719543) (10,0.7497454803) (11,0.7435339298) (12,0.7403290418) (13,0.7354335662) (14,0.7331098611) (15,0.7321921036) (16,0.7313266583) (17,0.7295019194) (18,0.7303431394) (19,0.7329689778) (20,0.7302437804)
};
\addplot[color=srModel9,dashed,line width=.85pt,mark=star,mark size=1.55pt,mark repeat=4] coordinates {
(1,0.8066442034) (2,0.8251473008) (3,0.8171372650) (4,0.8081258599) (5,0.7969742380) (6,0.7746942034) (7,0.7645833182) (8,0.7695936020) (9,0.7645243820) (10,0.7577751282) (11,0.7524692538) (12,0.7517029999) (13,0.7469935454) (14,0.7446444050) (15,0.7484863546) (16,0.7484189412) (17,0.7492945602) (18,0.7505688946) (19,0.7530871538) (20,0.7561834830)
};
\end{axis}
\begin{axis}[sr/base,width=150pt,height=93pt,at={(184pt,-124pt)},anchor=north west,
xmin=.5,xmax=20.5,ymin=0.6,ymax=1.01,xtick={1,5,10,15,20},ytick={.6,.7,.8,.9,1},xmajorgrids=false,
xlabel={Completed round},
]
\path[fill=black!9] (axis cs:.5,0.6) rectangle (axis cs:3.5,1.01);
\addplot[color=srModel0,solid,line width=.85pt,mark=*,mark size=1.55pt,mark repeat=4] coordinates {
(1,0.9892389111) (2,0.9778637611) (3,0.9700661074) (4,0.9652631889) (5,0.9562544511) (6,0.9392318481) (7,0.9395266905) (8,0.9336081014) (9,0.9299263926) (10,0.9289847133) (11,0.9271306798) (12,0.9220891500) (13,0.9173433538) (14,0.9141946921) (15,0.9117662037) (16,0.9093926104) (17,0.9075593144) (18,0.9050868907) (19,0.8994636854) (20,0.8959108022)
};
\addplot[color=srModel1,solid,line width=.85pt,mark=square*,mark size=1.55pt,mark repeat=4] coordinates {
(1,0.9274187444) (2,0.8988909499) (3,0.8834832221) (4,0.8774154861) (5,0.8733998355) (6,0.8524793518) (7,0.8416338936) (8,0.8379827222) (9,0.8321851061) (10,0.8268313277) (11,0.8256870252) (12,0.8187478037) (13,0.8175685376) (14,0.8171808079) (15,0.8139193118) (16,0.8172672791) (17,0.8137654679) (18,0.8071319370) (19,0.8023263877) (20,0.7978021083)
};
\addplot[color=srModel2,solid,line width=.85pt,mark=triangle*,mark size=1.55pt,mark repeat=4] coordinates {
(1,0.9883555556) (2,0.9741777778) (3,0.9661007407) (4,0.9626407639) (5,0.9577555556) (6,0.9476373593) (7,0.9513432921) (8,0.9433955764) (9,0.9372493049) (10,0.9349620856) (11,0.9343303535) (12,0.9298024435) (13,0.9278912829) (14,0.9257396000) (15,0.9250096563) (16,0.9219720153) (17,0.9191640824) (18,0.9159085623) (19,0.9112215515) (20,0.9091360378)
};
\addplot[color=srModel3,solid,line width=.85pt,mark=diamond*,mark size=1.55pt,mark repeat=4] coordinates {
(1,0.8747922222) (2,0.8651116667) (3,0.8580973037) (4,0.8539920000) (5,0.8412776956) (6,0.8199800167) (7,0.8183571111) (8,0.8174522833) (9,0.8091744556) (10,0.8004000533) (11,0.7932477667) (12,0.7890449806) (13,0.7840179402) (14,0.7795713873) (15,0.7728107667) (16,0.7642985042) (17,0.7554173092) (18,0.7476812377) (19,0.7422540573) (20,0.7367681500)
};
\addplot[color=srModel4,solid,line width=.85pt,mark=triangle,mark size=1.55pt,mark repeat=4] coordinates {
(1,0.9887188667) (2,0.9665804778) (3,0.9481324852) (4,0.9444940500) (5,0.9334185378) (6,0.9023746685) (7,0.9001450841) (8,0.8932462208) (9,0.8853803210) (10,0.8768934833) (11,0.8740899040) (12,0.8640859954) (13,0.8626816385) (14,0.8591306992) (15,0.8530336630) (16,0.8484327139) (17,0.8417555993) (18,0.8368779741) (19,0.8309592807) (20,0.8266807339)
};
\addplot[color=srModel5,dashed,line width=.85pt,mark=+,mark size=1.55pt,mark repeat=4] coordinates {
(1,0.9777632889) (2,0.9557527333) (3,0.9336107037) (4,0.9418744167) (5,0.9357215311) (6,0.9237488685) (7,0.9282887048) (8,0.9247523333) (9,0.9158561407) (10,0.9098257444) (11,0.9068913768) (12,0.9033666185) (13,0.8989369521) (14,0.8919014437) (15,0.8874036319) (16,0.8835180458) (17,0.8799331163) (18,0.8773986099) (19,0.8726589094) (20,0.8690419539)
};
\addplot[color=srModel6,dashed,line width=.85pt,mark=x,mark size=1.55pt,mark repeat=4] coordinates {
(1,0.9778922222) (2,0.9489338611) (3,0.9380730926) (4,0.9331654750) (5,0.9189617911) (6,0.9004359056) (7,0.9024651429) (8,0.8908947181) (9,0.8862350049) (10,0.8757581167) (11,0.8705246020) (12,0.8628917472) (13,0.8618475564) (14,0.8610869929) (15,0.8563717526) (16,0.8541586417) (17,0.8504422203) (18,0.8492016185) (19,0.8443361825) (20,0.8396916972)
};
\addplot[color=srModel7,dashed,line width=.85pt,mark=square,mark size=1.55pt,mark repeat=4] coordinates {
(1,0.9772100000) (2,0.9604327722) (3,0.9494240593) (4,0.9415324806) (5,0.9318293067) (6,0.9165749685) (7,0.9158385270) (8,0.9060767528) (9,0.8979300654) (10,0.8892227111) (11,0.8783999232) (12,0.8689999213) (13,0.8628391470) (14,0.8588115024) (15,0.8538465778) (16,0.8485344743) (17,0.8464485170) (18,0.8435042068) (19,0.8359009058) (20,0.8311941883)
};
\addplot[color=srModel8,dashed,line width=.85pt,mark=diamond,mark size=1.55pt,mark repeat=4] coordinates {
(1,0.9431044000) (2,0.9240183611) (3,0.9079633037) (4,0.9120449388) (5,0.9086967422) (6,0.8845276592) (7,0.8852473381) (8,0.8703638930) (9,0.8649757518) (10,0.8524282689) (11,0.8458731969) (12,0.8389749731) (13,0.8338025034) (14,0.8302119206) (15,0.8221145237) (16,0.8203667361) (17,0.8189210523) (18,0.8165113957) (19,0.8127320169) (20,0.8031554783)
};
\addplot[color=srModel9,dashed,line width=.85pt,mark=star,mark size=1.55pt,mark repeat=4] coordinates {
(1,0.9407866444) (2,0.9235816556) (3,0.9175028222) (4,0.9134068028) (5,0.9119137600) (6,0.8888749148) (7,0.8922152222) (8,0.8912931625) (9,0.8872007358) (10,0.8818062011) (11,0.8758876646) (12,0.8683144194) (13,0.8654903231) (14,0.8617007635) (15,0.8580879067) (16,0.8539300694) (17,0.8499272758) (18,0.8482139296) (19,0.8424966076) (20,0.8379668606)
};
\end{axis}
\end{tikzpicture}
\par\smallskip{\scriptsize\setlength{\tabcolsep}{5pt}\renewcommand{\arraystretch}{1.15}\begin{tabular}{ll}
\tikz[baseline=-.6ex]{\draw[srModel0,solid,line width=.8pt] (0,0)--(.42,0);\path (0.21,0) node[srModel0]{\pgfuseplotmark{*}};} GPT-6 Astra & \tikz[baseline=-.6ex]{\draw[srModel5,dashed,line width=.8pt] (0,0)--(.42,0);\path (0.21,0) node[srModel5]{\pgfuseplotmark{+}};} Gemini 3.8 Flash \\
\tikz[baseline=-.6ex]{\draw[srModel1,solid,line width=.8pt] (0,0)--(.42,0);\path (0.21,0) node[srModel1]{\pgfuseplotmark{square*}};} GPT-5.6 Terra & \tikz[baseline=-.6ex]{\draw[srModel6,dashed,line width=.8pt] (0,0)--(.42,0);\path (0.21,0) node[srModel6]{\pgfuseplotmark{x}};} Grok 4.6 \\
\tikz[baseline=-.6ex]{\draw[srModel2,solid,line width=.8pt] (0,0)--(.42,0);\path (0.21,0) node[srModel2]{\pgfuseplotmark{triangle*}};} Claude Fable 5.1 & \tikz[baseline=-.6ex]{\draw[srModel7,dashed,line width=.8pt] (0,0)--(.42,0);\path (0.21,0) node[srModel7]{\pgfuseplotmark{square}};} Grok 4.5 \\
\tikz[baseline=-.6ex]{\draw[srModel3,solid,line width=.8pt] (0,0)--(.42,0);\path (0.21,0) node[srModel3]{\pgfuseplotmark{diamond*}};} Claude Sonnet 5 & \tikz[baseline=-.6ex]{\draw[srModel8,dashed,line width=.8pt] (0,0)--(.42,0);\path (0.21,0) node[srModel8]{\pgfuseplotmark{diamond}};} Muse Spark 1.1 \\
\tikz[baseline=-.6ex]{\draw[srModel4,solid,line width=.8pt] (0,0)--(.42,0);\path (0.21,0) node[srModel4]{\pgfuseplotmark{triangle}};} Gemini 3.1 Pro Preview & \tikz[baseline=-.6ex]{\draw[srModel9,dashed,line width=.8pt] (0,0)--(.42,0);\path (0.21,0) node[srModel9]{\pgfuseplotmark{star}};} Muse Spark 1.2 \\
\end{tabular}}
\caption{\textbf{Score trajectories.} Cumulative means over all 10 games per domain, with matched axes. Shading marks the 3 uniform-weight warmup rounds, retained in all scores. Figure~\ref{fig:role-profiles} shows final-score intervals.}
\label{fig:trajectories}
\end{figure}

\subsection{Commitments and executed outputs}
\label{subsec:honesty}

Code-constrained games let us compare each private commitment with the program's executed output. Among $1{,}943$ valid proposals, $463$ commitments differ from that output ($23.8\%$; $95\%$ interval $[21.5\%,26.3\%]$). Figure~\ref{fig:execution-consistency} shows substantial variation across models. GPT-6 Astra and Claude Fable 5.1, the 2 highest-ranked models, have the lowest mismatch rates, $2/200$ ($1.0\%$) and $8/178$ ($4.5\%$), respectively. Grok 4.5 has the highest, $100/191$ ($52.4\%$).

The same figure compares mean proposer rewards on consistent and mismatching proposals. Among valid proposals, mismatches receive higher mean rewards for 8 of the 10 models. For Gemini 3.1 Pro Preview, the means are $0.517$ versus $0.028$, and for Claude Sonnet 5, $0.349$ versus $0.054$. Thus, mismatches can accompany strong separation, while the highest overall rankings coexist with low mismatch rates. An underlying association may arise through game dynamics, in which solvers respond differently to the observed commitment tendency of a proposer.

\begin{figure}[!htbp]
\centering
\begin{tikzpicture}
\begin{axis}[sr/forest,width=112pt,height=118pt,at={(0pt,0)},anchor=north west,
ymin=-.65,ymax=9.65,ytick={0,...,9},
yticklabels={{GPT-6 Astra},{Claude Fable 5.1},{Gemini 3.1 Pro Preview},{GPT-5.6 Terra},{Gemini 3.8 Flash},{Grok 4.6},{Muse Spark 1.1},{Muse Spark 1.2},{Claude Sonnet 5},{Grok 4.5}},
xmin=0,xmax=65,xtick={0,20,40,60},title={Mismatch rate},xlabel={Valid proposals (\%)},]
\path[fill=black!3] (axis cs:0,-0.45) rectangle (axis cs:65,0.45);
\path[fill=black!3] (axis cs:0,1.55) rectangle (axis cs:65,2.45);
\path[fill=black!3] (axis cs:0,3.55) rectangle (axis cs:65,4.45);
\path[fill=black!3] (axis cs:0,5.55) rectangle (axis cs:65,6.45);
\path[fill=black!3] (axis cs:0,7.55) rectangle (axis cs:65,8.45);
\addplot[only marks,mark=*,mark size=2pt,color=srJoint,error bars/.cd,x dir=both,x explicit,error bar style={line width=.75pt},error mark options={rotate=90,mark size=1.7pt}]
table[x=x,y=y,x error minus=em,x error plus=ep] {
x y em ep
1.0000000000 0 1.0000000000 1.5000000000
4.4943820225 1 3.3770635867 4.7541728908
9.1370558376 2 3.5532994924 3.6834569829
18.3673469388 3 6.9090136054 7.1326530612
24.7422680412 4 6.6075530153 7.7319587629
28.5714285714 5 4.4508255564 5.2747252747
29.9492385787 6 10.1522842640 11.0507614213
33.5025380711 7 10.6598984772 11.6751269036
35.5329949239 8 6.7451161360 5.7935356884
52.3560209424 9 8.9005235602 8.5814790576
};
\node[anchor=west,font=\scriptsize,text=black!70] at (axis cs:67,0) {2/200};
\node[anchor=west,font=\scriptsize,text=black!70] at (axis cs:67,1) {8/178};
\node[anchor=west,font=\scriptsize,text=black!70] at (axis cs:67,2) {18/197};
\node[anchor=west,font=\scriptsize,text=black!70] at (axis cs:67,3) {36/196};
\node[anchor=west,font=\scriptsize,text=black!70] at (axis cs:67,4) {48/194};
\node[anchor=west,font=\scriptsize,text=black!70] at (axis cs:67,5) {56/196};
\node[anchor=west,font=\scriptsize,text=black!70] at (axis cs:67,6) {59/197};
\node[anchor=west,font=\scriptsize,text=black!70] at (axis cs:67,7) {66/197};
\node[anchor=west,font=\scriptsize,text=black!70] at (axis cs:67,8) {70/197};
\node[anchor=west,font=\scriptsize,text=black!70] at (axis cs:67,9) {100/191};
\end{axis}
\begin{axis}[sr/forest,width=103pt,height=118pt,at={(166pt,0)},anchor=north west,
ymin=-.65,ymax=9.65,ytick={0,...,9},
yticklabels={{},{},{},{},{},{},{},{},{},{}},
xmin=0,xmax=.70,xtick={0,.2,.4,.6},title={Conditional reward},xlabel={Mean proposer reward $P$},]
\path[fill=black!3] (axis cs:0,-0.45) rectangle (axis cs:.70,0.45);
\path[fill=black!3] (axis cs:0,1.55) rectangle (axis cs:.70,2.45);
\path[fill=black!3] (axis cs:0,3.55) rectangle (axis cs:.70,4.45);
\path[fill=black!3] (axis cs:0,5.55) rectangle (axis cs:.70,6.45);
\path[fill=black!3] (axis cs:0,7.55) rectangle (axis cs:.70,8.45);
\addplot[black!40,no marks,line width=.8pt] coordinates {(0.6287390312,0) (0.6118405041,0)};
\addplot[black!40,no marks,line width=.8pt] coordinates {(0.5762881614,1) (0.5307210519,1)};
\addplot[black!40,no marks,line width=.8pt] coordinates {(0.0280787352,2) (0.5168925531,2)};
\addplot[black!40,no marks,line width=.8pt] coordinates {(0.3851759898,3) (0.4200043457,3)};
\addplot[black!40,no marks,line width=.8pt] coordinates {(0.2623179506,4) (0.4684085176,4)};
\addplot[black!40,no marks,line width=.8pt] coordinates {(0.1981368457,5) (0.5291159420,5)};
\addplot[black!40,no marks,line width=.8pt] coordinates {(0.1804318916,6) (0.4341262793,6)};
\addplot[black!40,no marks,line width=.8pt] coordinates {(0.1466618569,7) (0.4701874376,7)};
\addplot[black!40,no marks,line width=.8pt] coordinates {(0.0540492404,8) (0.3490861868,8)};
\addplot[black!40,no marks,line width=.8pt] coordinates {(0.1994628979,9) (0.4174749895,9)};
\addplot[only marks,color=srCode,mark=*,mark size=2pt] coordinates {
(0.6287390312,0) (0.5762881614,1) (0.0280787352,2) (0.3851759898,3) (0.2623179506,4) (0.1981368457,5) (0.1804318916,6) (0.1466618569,7) (0.0540492404,8) (0.1994628979,9)
};
\addplot[only marks,color=srOpen,mark=diamond*,mark size=2pt] coordinates {
(0.6118405041,0) (0.5307210519,1) (0.5168925531,2) (0.4200043457,3) (0.4684085176,4) (0.5291159420,5) (0.4341262793,6) (0.4701874376,7) (0.3490861868,8) (0.4174749895,9)
};
\end{axis}
\end{tikzpicture}
\par\smallskip{\footnotesize\textcolor{srCode}{$\bullet$ Consistent}\qquad\textcolor{srOpen}{$\blacklozenge$ Mismatching}}
\caption{\textbf{Commitments versus executed outputs.} The left panel shows code-constrained mismatch rates with $95\%$ game-bootstrap intervals and counts (mismatches/valid proposals). The right panel shows mean rewards pooled over valid proposals.}
\label{fig:execution-consistency}
\end{figure}

\subsection{Anonymous author ablation}
\label{subsec:author-disclosure}

The default protocol lets solvers use a pseudonymous proposer's earlier behavior when predicting its next commitment. To examine this information channel, we run an anonymous author ablation of 5 open-ended games in which solvers are not told the current proposer's pseudonym. We compare them with the first 10 rounds of the 5 standard games, in which the proposer is disclosed.

Figure~\ref{fig:author-ablation} reports paired differences, anonymous minus disclosed, at this common 10-round horizon. Averaged equally over the roster, anonymity changes proposer reward by $0.026$ [$-0.066$, $0.118$], solver skill by $-0.025$ [$-0.088$, $0.044$], and joint score by $0.001$ [$-0.016$, $0.018$]. The point estimates move the 2 roles in opposite directions and largely cancel in the joint score. The intervals use all $5^5$ ordered resamples of the 5 matched game pairs.

\begin{figure}[!htbp]
\centering
\begingroup
\definecolor{srAblationPoint}{HTML}{75488F}
\definecolor{srAblationRoster}{HTML}{182E42}
\begin{tikzpicture}
\begin{groupplot}[
 group style={group size=3 by 1,horizontal sep=11pt},
 width=.19\linewidth,height=140pt,scale only axis,
 xmin=-.28,xmax=.32,ymin=-.65,ymax=10.65,y dir=reverse,
 xtick={-.2,0,.2},ytick={0,...,10},
 scaled ticks=false,tick label style={font=\scriptsize},
 tick align=outside,ytick style={draw=none},
 title style={font=\small},axis lines*=left,
 axis line style={black!35},xtick style={black!35},
 xmajorgrids,grid style={black!8},clip=false,
 every axis plot/.append style={line width=.7pt},
]

\nextgroupplot[title={$\Delta P$},yticklabels={{GPT-6 Astra},{GPT-5.6 Terra},{Claude Fable 5.1},{Claude Sonnet 5},{Gemini 3.1 Pro Preview},{Gemini 3.8 Flash},{Grok 4.6},{Grok 4.5},{Muse Spark 1.1},{Muse Spark 1.2},{Equal-weight roster}},yticklabel style={font=\scriptsize,align=right}]
\path[fill=black!3] (axis cs:-.28,-0.42) rectangle (axis cs:.32,0.42);
\path[fill=black!3] (axis cs:-.28,1.58) rectangle (axis cs:.32,2.42);
\path[fill=black!3] (axis cs:-.28,3.58) rectangle (axis cs:.32,4.42);
\path[fill=black!3] (axis cs:-.28,5.58) rectangle (axis cs:.32,6.42);
\path[fill=black!3] (axis cs:-.28,7.58) rectangle (axis cs:.32,8.42);
\path[fill=black!3] (axis cs:-.28,9.58) rectangle (axis cs:.32,10.42);
\addplot[black!45,densely dashed,no marks] coordinates {(0,-.5) (0,10.5)};
\addplot[black!25,no marks] coordinates {(-.28,9.5) (.32,9.5)};
\addplot[only marks,mark=*,mark size=2pt,srAblationPoint,error bars/.cd,x dir=both,x explicit,error bar style={line width=.7pt},error mark options={rotate=90,mark size=1.8pt}]
table[x=x,y=y,x error minus=minus,x error plus=plus,row sep=\\] {x y minus plus\\
0.019291526236 0 0.234826825291 0.234826825291\\ 
0.139890539541 1 0.150845536936 0.150845536936\\ 
0.063894845749 2 0.130420700663 0.158421227505\\ 
0.017914092303 3 0.022762637115 0.033971594547\\ 
-0.003427979029 4 0.036764998687 0.041115114266\\ 
0.006201662705 5 0.062230596566 0.089279403496\\ 
0.002291207254 6 0.151262034539 0.147457883419\\ 
-0.033141234755 7 0.091369514419 0.096406430670\\ 
-0.038338310559 8 0.088027807348 0.088027807348\\ 
0.087372836740 9 0.118612525635 0.114635264489\\ 
};
\addplot[only marks,mark=diamond*,mark size=2.7pt,srAblationRoster,error bars/.cd,x dir=both,x explicit,error bar style={line width=.7pt},error mark options={rotate=90,mark size=1.8pt}]
table[x=x,y=y,x error minus=minus,x error plus=plus,row sep=\\] {x y minus plus\\
0.026194918618 10 0.091748974951 0.091748974951\\ 
};
\nextgroupplot[title={$\Delta S$},yticklabels={},yticklabel style={font=\scriptsize,align=right}]
\path[fill=black!3] (axis cs:-.28,-0.42) rectangle (axis cs:.32,0.42);
\path[fill=black!3] (axis cs:-.28,1.58) rectangle (axis cs:.32,2.42);
\path[fill=black!3] (axis cs:-.28,3.58) rectangle (axis cs:.32,4.42);
\path[fill=black!3] (axis cs:-.28,5.58) rectangle (axis cs:.32,6.42);
\path[fill=black!3] (axis cs:-.28,7.58) rectangle (axis cs:.32,8.42);
\path[fill=black!3] (axis cs:-.28,9.58) rectangle (axis cs:.32,10.42);
\addplot[black!45,densely dashed,no marks] coordinates {(0,-.5) (0,10.5)};
\addplot[black!25,no marks] coordinates {(-.28,9.5) (.32,9.5)};
\addplot[only marks,mark=*,mark size=2pt,srAblationPoint,error bars/.cd,x dir=both,x explicit,error bar style={line width=.7pt},error mark options={rotate=90,mark size=1.8pt}]
table[x=x,y=y,x error minus=minus,x error plus=plus,row sep=\\] {x y minus plus\\
-0.015855084444 0 0.051267308889 0.051267308889\\ 
-0.016598993422 1 0.084191440111 0.084191440111\\ 
-0.023090780000 2 0.044687264444 0.042421193333\\ 
-0.030164991111 3 0.091021893333 0.075930695556\\ 
-0.003439062244 4 0.072142435533 0.084507195556\\ 
-0.031360224467 5 0.049910937756 0.063152059956\\ 
-0.034722011111 6 0.054808362222 0.066159920000\\ 
-0.041461800000 7 0.080989460000 0.096437791111\\ 
-0.031985280000 8 0.076834562222 0.076834562222\\ 
-0.019007004444 9 0.077776326667 0.078954984444\\ 
};
\addplot[only marks,mark=diamond*,mark size=2.7pt,srAblationRoster,error bars/.cd,x dir=both,x explicit,error bar style={line width=.7pt},error mark options={rotate=90,mark size=1.8pt}]
table[x=x,y=y,x error minus=minus,x error plus=plus,row sep=\\] {x y minus plus\\
-0.024768523124 10 0.063284208667 0.068534365569\\ 
};
\nextgroupplot[title={$\Delta J$},yticklabels={},yticklabel style={font=\scriptsize,align=right}]
\path[fill=black!3] (axis cs:-.28,-0.42) rectangle (axis cs:.32,0.42);
\path[fill=black!3] (axis cs:-.28,1.58) rectangle (axis cs:.32,2.42);
\path[fill=black!3] (axis cs:-.28,3.58) rectangle (axis cs:.32,4.42);
\path[fill=black!3] (axis cs:-.28,5.58) rectangle (axis cs:.32,6.42);
\path[fill=black!3] (axis cs:-.28,7.58) rectangle (axis cs:.32,8.42);
\path[fill=black!3] (axis cs:-.28,9.58) rectangle (axis cs:.32,10.42);
\addplot[black!45,densely dashed,no marks] coordinates {(0,-.5) (0,10.5)};
\addplot[black!25,no marks] coordinates {(-.28,9.5) (.32,9.5)};
\addplot[only marks,mark=*,mark size=2pt,srAblationPoint,error bars/.cd,x dir=both,x explicit,error bar style={line width=.7pt},error mark options={rotate=90,mark size=1.8pt}]
table[x=x,y=y,x error minus=minus,x error plus=plus,row sep=\\] {x y minus plus\\
0.001718220896 0 0.096613531535 0.096613531535\\ 
0.061645773059 1 0.088942667949 0.088942667949\\ 
0.020402032874 2 0.048749524776 0.062228903893\\ 
-0.006125449404 3 0.051858126287 0.037515570755\\ 
-0.003433520637 4 0.029339998618 0.032855545950\\ 
-0.012579280881 5 0.036890511464 0.037232810266\\ 
-0.016215401929 6 0.047126768376 0.050629438325\\ 
-0.037301517378 7 0.028710519801 0.024020635195\\ 
-0.035161795279 8 0.035045386059 0.038832799508\\ 
0.034182916148 9 0.050965862370 0.045729264933\\ 
};
\addplot[only marks,mark=diamond*,mark size=2.7pt,srAblationRoster,error bars/.cd,x dir=both,x explicit,error bar style={line width=.7pt},error mark options={rotate=90,mark size=1.8pt}]
table[x=x,y=y,x error minus=minus,x error plus=plus,row sep=\\] {x y minus plus\\
0.000713197747 10 0.016906111404 0.016906111404\\ 
};
\end{groupplot}
\node[anchor=north,font=\small] at ([yshift=-23pt]group c2r1.south) {Anonymous $-$ disclosed};
\end{tikzpicture}
\endgroup
\caption{\textbf{Anonymous author ablation.} Differences between 5 matched pairs of open-ended games, all measured after 10 rounds. Bars show pointwise 95\% paired game-bootstrap intervals.}
\label{fig:author-ablation}
\end{figure}

\end{document}